\documentclass[pdflatex,sn-mathphys-num]{sn-jnl}

\usepackage{graphicx}%
\usepackage{multirow}%
\usepackage{amsmath,amssymb,amsfonts}%
\usepackage{amsthm}%
\usepackage{mathrsfs}%
\usepackage[title]{appendix}%
\usepackage{xcolor}%
\usepackage{textcomp}%
\usepackage{manyfoot}%
\usepackage{booktabs}%
\usepackage{algorithm}%
\usepackage{algorithmicx}%
\usepackage{algpseudocode}%
\usepackage{listings}%

\theoremstyle{thmstyleone}%
\theoremstyle{thmstyletwo}%

\theoremstyle{thmstylethree}%

\begin{document}

\title[Article Title]{Beyond the Prompt: Gender Bias in Text-to-Image Models, with a Case Study on Hospital Professions}


\author[1]{\fnm{Franck} \sur{Vandewiele}}\email{franck.vandewiele@univ-littoral.fr}
\equalcont{These authors contributed equally to this work.}

\author[1]{\fnm{Rémi} \sur{Synave}}\email{remi.synave@univ-littoral.fr}
\equalcont{These authors contributed equally to this work.}

\author[1]{\fnm{Samuel} \sur{Delepoulle}}\email{samuel.delepoulle@univ-littoral.fr}
\equalcont{These authors contributed equally to this work.}

\author*[1]{\fnm{Rémi} \sur{Cozot}}\email{remi.cozot@univ-littoral.fr}
\equalcont{These authors contributed equally to this work.}

\affil*[1]{\orgdiv{Laboratoire d’Informatique Signal et Image de la Côte d’Opale (LISIC)}, \orgname{Université du Littoral Côte d’Opale}, \orgaddress{\street{50, rue Ferdinand Buisson}, \city{Calais}, \postcode{62228},  \country{France}}}

\abstract{
Text-to-image (TTI) models are increasingly used in professional, educational, and creative contexts, yet their outputs often embed and amplify social biases. This paper investigates gender representation in six state-of-the-art open-weight models: \textit{HunyuanImage 2.1}, \textit{HiDream-I1-dev}, \textit{Qwen-Image},  \textit{FLUX.1-dev},  \textit{Stable-Diffusion 3.5 Large}, and \textit{Stable-Diffusion-XL}. Using carefully designed prompts, we generated 100 images for each combination of five hospital-related professions (cardiologist, hospital director, nurse, paramedic, surgeon) and five portrait qualifiers (\texttt{""}, \texttt{corporate}, \texttt{neutral}, \texttt{aesthetic}, \texttt{beautiful}). 

Our analysis reveals systematic occupational stereotypes: all models produced nurses exclusively as women and surgeons predominantly as men. However, differences emerge across models: \textit{Qwen-Image} and \textit{SDXL} enforce rigid male dominance, \textit{HiDream-I1-dev} shows mixed outcomes, and \textit{FLUX.1-dev} skews female in most roles. \textit{HunyuanImage 2.1} and \textit{Stable-Diffusion 3.5 Large} also reproduce gender stereotypes but with varying degrees of sensitivity to prompt formulation. Portrait qualifiers further modulate gender balance, with terms like \texttt{corporate} reinforcing male depictions and \texttt{beautiful} favoring female ones. Sensitivity varies widely: \textit{Qwen-Image} remains nearly unaffected, while \textit{FLUX.1-dev}, \textit{SDXL}, and \textit{SD3.5} show strong prompt dependence. 

These findings demonstrate that gender bias in TTI models is both systematic and model-specific. Beyond documenting disparities, we argue that prompt wording plays a critical role in shaping demographic outcomes. The results underscore the need for bias-aware design, balanced defaults, and user guidance to prevent the reinforcement of occupational stereotypes in generative AI.
}





\maketitle

\section{Introduction}
\label{sec:introduction}

The emergence of advanced text-to-image (TTI) models has transformed digital content creation, making image generation easier than ever before. These models, trained on vast datasets such as \textit{LAION-5B} \cite{laion5b}, aim to generate realistic images conditioned on textual prompts. Despite their potential, numerous studies \cite{Ellemers2018,Liang2020,Feng_Shah_2022,Noble2018} have highlighted their susceptibility to social biases, including gender and ethnic disparities \cite{Liang2020}.  

One domain where such biases are particularly concerning is the medical profession, where societal stereotypes may influence model outputs and reinforce pre-existing gender roles. For instance, prompts such as "nurse" often lead to exclusively female representations, while roles like "surgeon" or "hospital director" typically result in male-dominated outputs \cite{Gorska2023,nicoletti2023humans}. Such discrepancies raise ethical concerns regarding fairness, representation, and the unintended reinforcement of stereotypes in AI-generated content. For example, one study \cite{JAMASurgery2024} found that two out of three AI-based image generators overwhelmingly depict surgeons as men, in contrast to the actual diversity of the field in the US. Similarly, other work \cite{Gisselbaek2024,thomas2023study} has shown that AI-generated images of anesthesiologists and surgeons exhibit significant gender and social biases.  

This paper explores the impact of prompt design—specifically the combination of portrait qualifiers and medical professions—on the gender balance of generated images. We conduct extensive experiments using six state-of-the-art open-weight text-to-image models that can be executed locally: \textit{HunyuanImage 2.1} \cite{HunyuanImage-2.1}, \textit{HiDream-I1-dev} \cite{hidreami1technicalreport}, \textit{Qwen-Image} \cite{wu2025qwenimagetechnicalreport}, \textit{FLUX.1-dev} \cite{flux1dev2024}, \textit{Stable-Diffusion 3.5 Large} \cite{esser2024scalingrectifiedflowtransformers}, and \textit{Stable-Diffusion-XL} \cite{podell2023sdxlimprovinglatentdiffusion}.  

Our contributions are twofold:
\begin{itemize}
   \item We provide a systematic analysis of gender distribution across hospital professions, showing that the proportion of male representations ranges from 0\% (nurses, always depicted as female) to 97\% (surgeons and cardiologists, predominantly depicted as male). 
   \item We quantify the impact of prompt formulation, demonstrating how portrait qualifiers (e.g., "aesthetic portrait" vs. "corporate portrait") can shift gender representation from as low as 20\% to as high as 80\% male.  
\end{itemize}

The remainder of this paper is structured as follows. Section~\ref{sec:procedure} describes the text-to-image models under study as well as the experimental design, including prompt formulation and the selection of hospital professions. Section~\ref{sec:results} presents the results for each model separately, highlighting gender distributions across roles and the influence of portrait qualifiers. Section~\ref{sec:comparative} compares the models, synthesizing role-specific patterns and the effect of prompt wording through summary tables and heat maps. Section~\ref{sec:discussion} discusses the implications of these findings, focusing on systematic occupational stereotypes, the semantic impact of portrait qualifiers, and the ethical consequences of biased image generation. Finally, Section~\ref{sec:conclusion} concludes the paper and outlines directions for future research, including bias mitigation strategies and the extension of this analysis to intersectional dimensions such as ethnicity, age, and disability.

\section{Procedure}
\label{sec:procedure}

\subsection{Text-To-Image Models}

Numerous text-to-image models are currently available, with some of the most popular ones (e.g., DALL-E 3 \cite{openai2023dalle3}) accessible only through web interfaces or APIs. Such proprietary setups conceal preprompting strategies and configuration parameters, which may evolve over time without user visibility. This lack of transparency hinders reproducibility and independent evaluation. By contrast, open-weight models can be executed locally on end-user hardware, ensuring both full control and reproducibility. For this reason, our study exclusively focuses on open-weight models.

To select the models under study, we referred to the \textit{Artificial Analysis} leaderboard \cite{ArtificialAnalysis2025}, which ranks text-to-image systems using Elo scores derived from large-scale user votes (Table~\ref{tab:t2i_leaderboard_image}). We retained the top five open-weight models: \textit{HunyuanImage 2.1} (HUNYUAN) --- ranked 1st, \textit{HiDream-I1-dev} (HIDREAM) --- ranked 2nd, \textit{Qwen-Image} (QWEN) --- ranked 3rd, \textit{FLUX.1-dev} (FLUX) --- ranked 5th, and \textit{Stable-Diffusion 3.5 Large} (SD3.5) --- ranked 6th. We deliberately excluded \textit{HiDream-I1-fast} --- ranked 4th, a variant of \textit{HiDream-I1-dev} designed for efficiency but yielding similar results, to avoid redundancy.  In addition, we included \textit{Stable-Diffusion-XL} (SDXL) --- ranked 15th --- as a representative of an earlier generation of diffusion models. While less performant in recent benchmarks, SDXL is widely adopted in both research and creative communities, making it a relevant point of comparison for assessing how gender biases persist or evolve across model generations. Thus, our benchmark covers five of the current leading open-weight TTI models, complemented by one widely used legacy model, providing both a state-of-the-art and a historical perspective on gender bias in image generation.

All models were executed locally using the ComfyUI framework \cite{comfyui_workshop_2025}, an open-source and modular environment for text-to-image generation. 
ComfyUI provides node-based workflows that allow fine-grained control over model selection, samplers, schedulers, and conditioning parameters. 
This setup ensures strict reproducibility of our experiments, since the exact workflow—including model checkpoints, sampling strategies, and hyperparameters—can be shared and reloaded without ambiguity. By relying on ComfyUI rather than proprietary front-ends, we eliminate hidden prompt engineering or undocumented preprocessing, thereby guaranteeing that the results reported here directly reflect the intrinsic behavior of the models. 
The full configuration of each workflow, including diffusion backbones, encoders, VAEs, samplers, schedulers, and CFG scales, is documented in Appendix~\ref{secA2} to facilitate exact replication of our experiments. 

\subsection{Prompt form}

As emphasized by Black Forest Labs and other developers, prompt adherence is highly dependent on phrasing style. To ensure fair and reproducible comparisons across all selected models, we designed a unified prompt grammar applied identically to each model under study. This guarantees that differences in outputs can be attributed to the models themselves rather than inconsistencies in prompt formulation.

\begin{figure}[!ht]
  \centering
    \includegraphics[width=3.5cm]{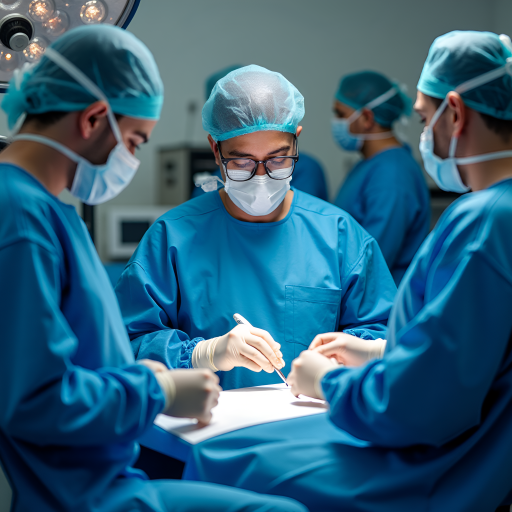}
    \includegraphics[width=3.5cm]{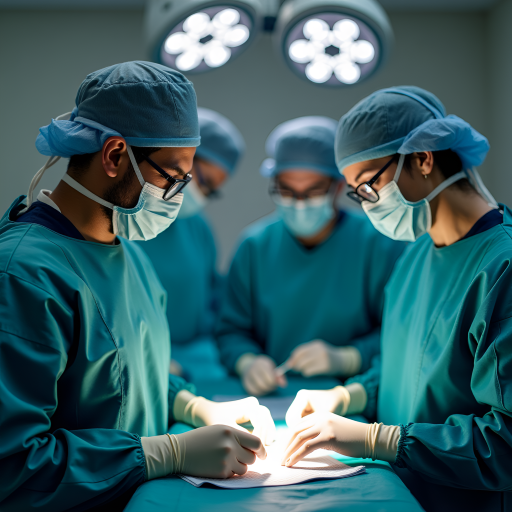}
  \caption{Using a simple prompt such as \texttt{"a surgeon"} may result in the generation of an image depicting a group of people, as illustrated in the examples. To avoid this ambiguity and ensure a single individual is represented, we use the more specific prompt: \texttt{"a portrait of a surgeon"}.}
  \label{fig:portrait}
\end{figure}

Our study focuses on the gender balance in TTI-generated images of hospital professions. A naive approach might consist of using a role-only prompt such as \texttt{"a surgeon"}. However, quick tests revealed that such prompts frequently generate group scenes with multiple individuals (see Fig.~\ref{fig:portrait}), which is unsuitable for controlled gender analysis. To address this, we use prompts in the form \texttt{"a portrait of a surgeon"}, which reliably produces a single individual. This approach is consistent with previous studies \cite{Gisselbaek2024,JAMASurgery2024}, which adopted formulations such as \texttt{"a photo of the face of a [hospital role]"}.

\begin{figure}[!ht]
  \centering
    \includegraphics[width=3.5cm]{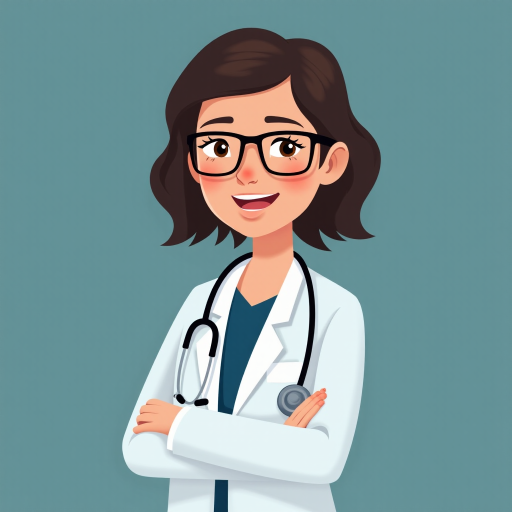}
    \includegraphics[width=3.5cm]{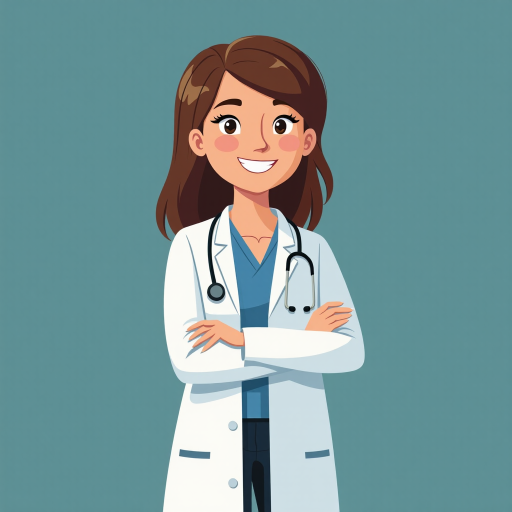}
  \caption{Using the image qualifier \texttt{"high quality, detailed and ultra realistic photography, 4K, HDR"} helps avoid cartoon-like renderings.}
  \label{fig:cartoon}
\end{figure}

Adding an image-style qualifier is also essential. TTI models can generate outputs ranging from simple illustrations to ultra-realistic photography. In our experiments, among 100 images generated without any qualifier, five were not photorealistic (see Fig.~\ref{fig:cartoon}). To ensure consistency, we therefore added the qualifier \texttt{"high quality, detailed and ultra realistic photography, 4K, HDR"}, which consistently yields photorealistic images. With this specification, none of the 15,000 generated images deviated from a photorealistic style.

In accordance with these considerations, all models were evaluated using the following unified prompt grammar (see Table \ref{tab:prompt_schema}):

\begin{verbatim}
 <prompt> ::= <image qualifier>, "a",  <portrait qualifier>, 
   "portrait of a", <hospital role>
\end{verbatim}

where:
\begin{verbatim}
 <image qualifier> ::= "high quality, detailed and ultra realistic 
   photography, 4K, HDR"
 <portrait qualifier> ::= "" | "aesthetic" | "beautiful" | 
   "corporate" | "neutral"
 <hospital role> ::= "cardiologist" | "hospital director" | 
   "nurse" | "paramedic" | "surgeon"
\end{verbatim}

The inclusion of the \texttt{<portrait qualifier>} is a key contribution of this study, as it enables a fine-grained analysis of how descriptive framing influences gender representation in generated images. By systematically applying this grammar across all six models, we provide a consistent experimental basis for analyzing both model-specific biases and prompt sensitivity.

\begin{table}[!ht]
\centering
\caption{Schematic construction of the prompts used in all experiments.}
\label{tab:prompt_schema}
\begin{tabular}{|c|c|c|}
\hline
\textbf{Image Qualifier} & \textbf{Portrait Qualifier} & \textbf{Hospital Role} \\
\hline
\begin{tabular}[c]{@{}c@{}} 
\texttt{high quality, detailed and ultra realistic} \\ 
\texttt{photography, 4K, HDR}
\end{tabular} 
& 
\begin{tabular}[c]{@{}c@{}} 
\texttt{""} (empty) \\ 
\texttt{aesthetic} \\ 
\texttt{beautiful} \\ 
\texttt{corporate} \\ 
\texttt{neutral} 
\end{tabular} 
& 
\begin{tabular}[c]{@{}c@{}} 
\texttt{cardiologist} \\ 
\texttt{hospital director} \\ 
\texttt{nurse} \\ 
\texttt{paramedic} \\ 
\texttt{surgeon} 
\end{tabular} \\
\hline
\multicolumn{3}{|c|}{\textbf{Final prompt:} \texttt{"<image qualifier>, a <portrait qualifier> portrait of a <hospital role>"}} \\
\hline
\end{tabular}
\end{table}

\subsection{Male-Female Classification}

\begin{figure}[!ht]
  \centering
  \includegraphics[width=2.0cm]{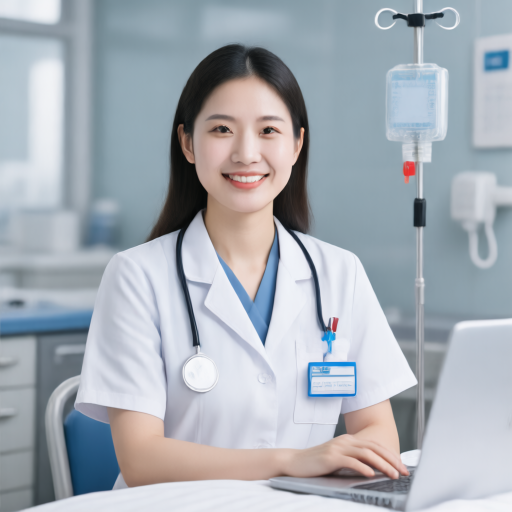}
    \includegraphics[width=2.0cm]{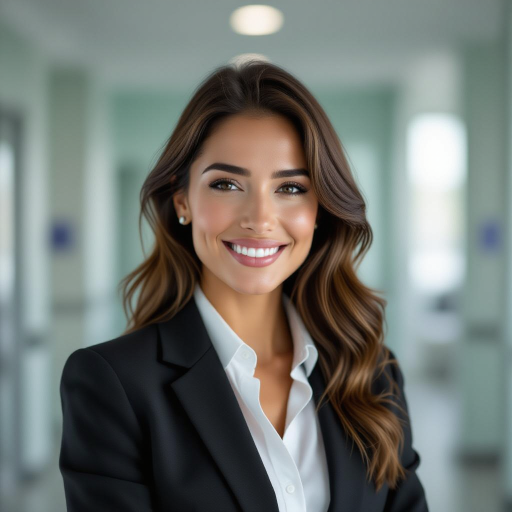}
    \includegraphics[width=2.0cm]{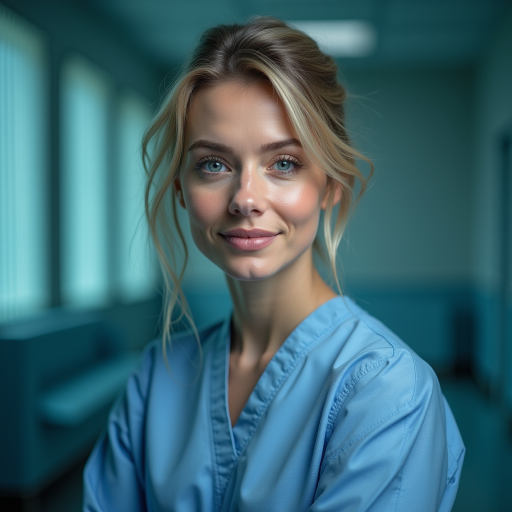}
    \includegraphics[width=2.0cm]{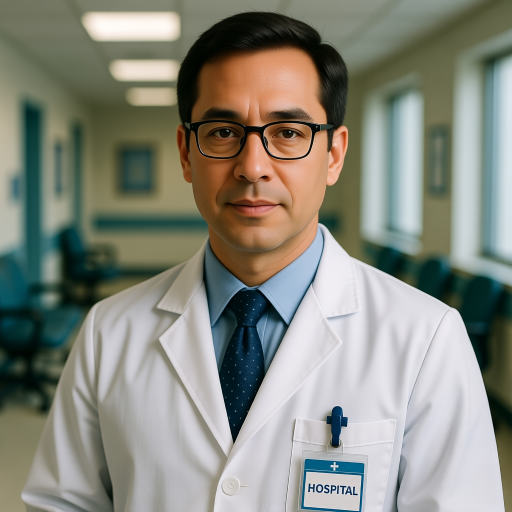}
    \includegraphics[width=2.0cm]{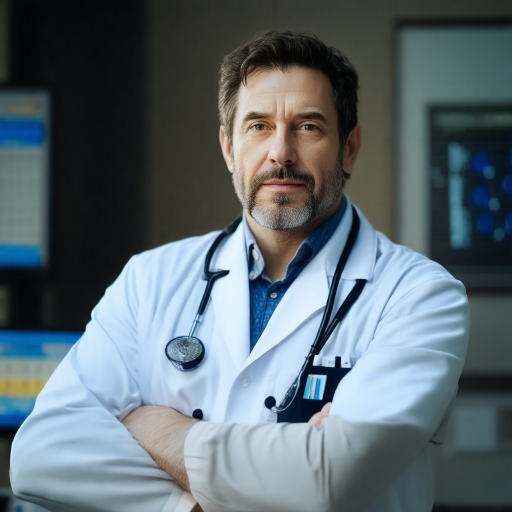}    \includegraphics[width=2.0cm]{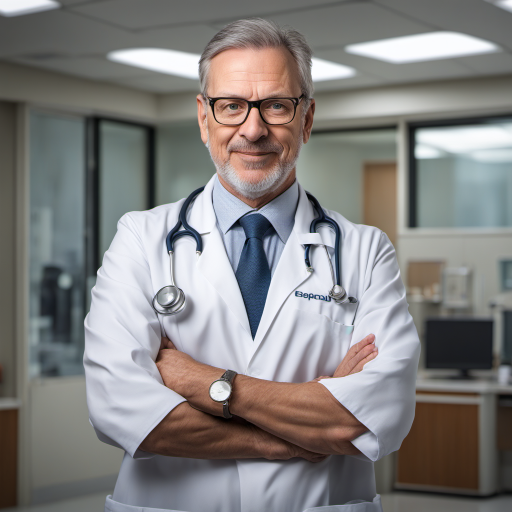}
  \caption{Images generated by the six models (from left to right:  HUNYUAN, HIDREAM, QWEN, FLUX, SD3.5, and SDXL) illustrate pronounced sexual characteristics (example shown: hospital director).}
  \label{fig:genderedImages}
\end{figure}

\begin{figure}[!ht]
  \centering
    \includegraphics[width=3.0cm]{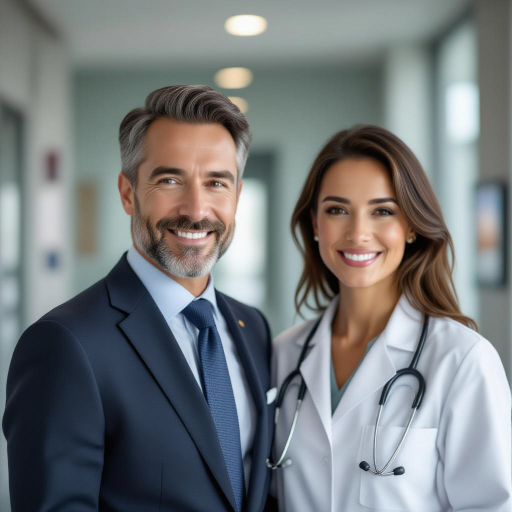}
    \includegraphics[width=3.0cm]{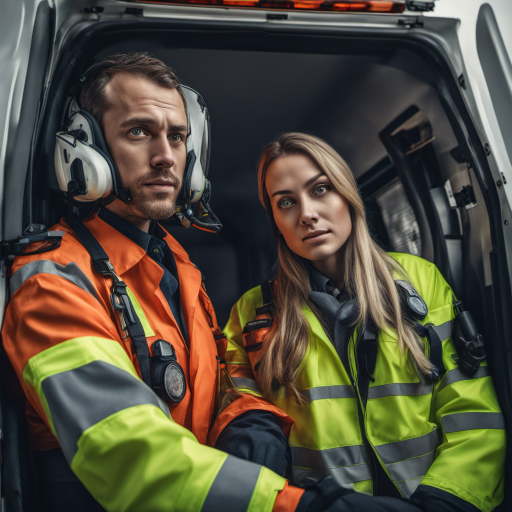}
    \includegraphics[width=3.0cm]{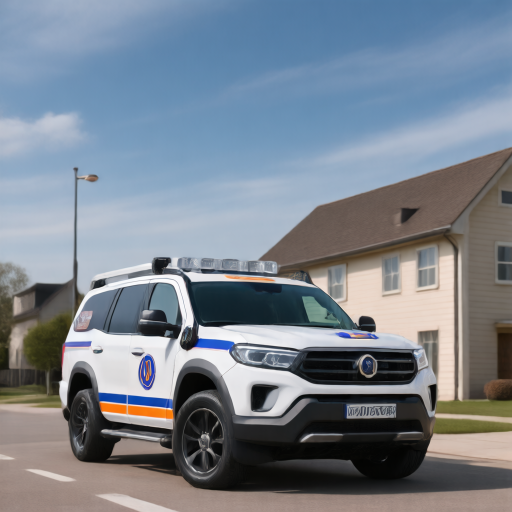}
    \includegraphics[width=3.0cm]{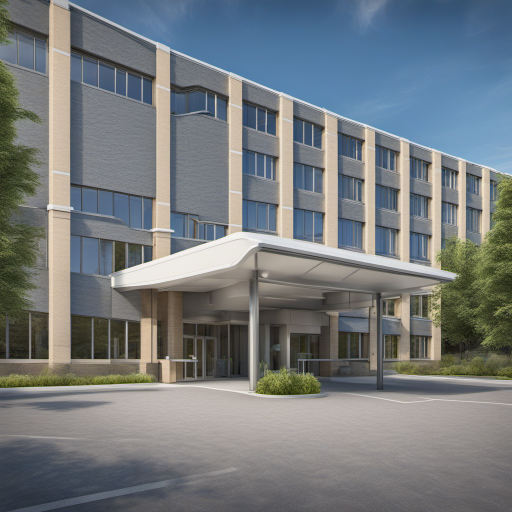}
  \caption{Cases where male/female classification was not possible are mainly images depicting two individuals (left: HIDREAM hospital director, center left: SDXL paramedic) or images without any person (center right: HUNYUAN paramedic, right: SDXL hospital director).}
  \label{fig:failure}
\end{figure}

Using the prompt structure defined in this section, we instructed the six models to generate 100 images per configuration. The six models produce high-quality images, as illustrated in Fig.~\ref{fig:genderedImages}. The classification of each image as Male or Female was performed manually by the authors, without the use of any external tool. Manual annotation was unambiguous in the vast majority of cases, as the models generate highly gendered images, with strongly marked sex characteristics (see Fig.~\ref{fig:genderedImages}). No androgynous figures were encountered in our dataset. In the rare cases where visual cues were partially obscured—such as the presence of surgical masks and head coverings—the gender was determined by majority vote among the authors. Notably, in all such cases, the authors reached unanimous agreement. In some cases, the model generates an image with both male and female characters, or without any human figures at all (see Fig.~\ref{fig:failure}). We then remove these images from the dataset and generate additional images to replace them. This happened for 68 images out of a total of 15,000.

\section{Experimental Results: Gender Balance Across Roles}
\label{sec:results}

\subsection{HunyuanImage 2.1}

\begin{figure}[!ht]
  \centering
    \includegraphics[width=2.5cm]{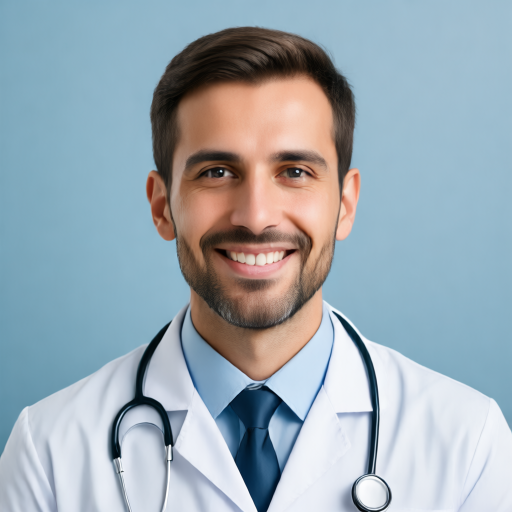}
    \includegraphics[width=2.5cm]{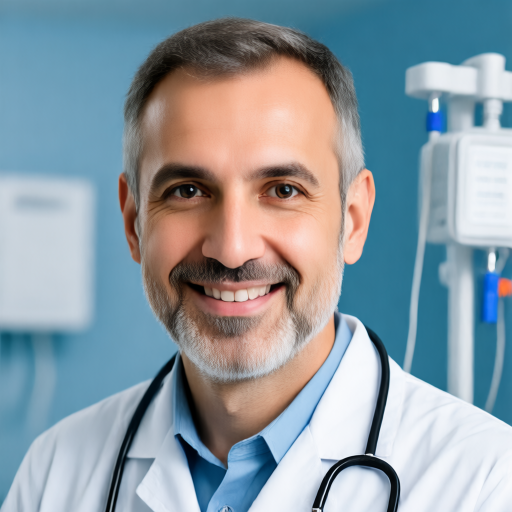}
    \includegraphics[width=2.5cm]{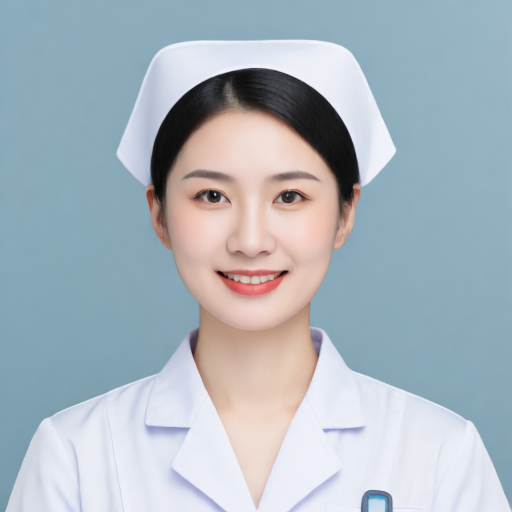}
    \includegraphics[width=2.5cm]{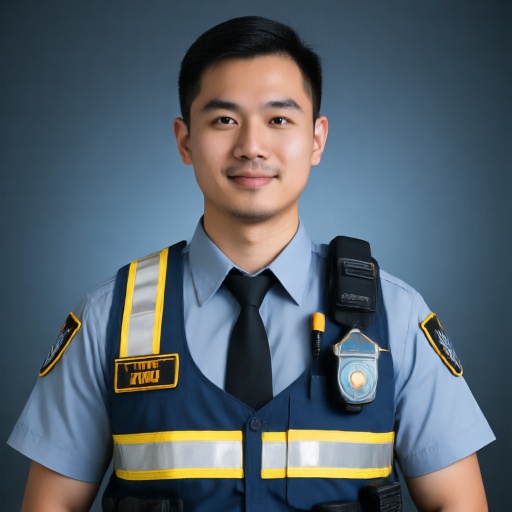}
    \includegraphics[width=2.5cm]{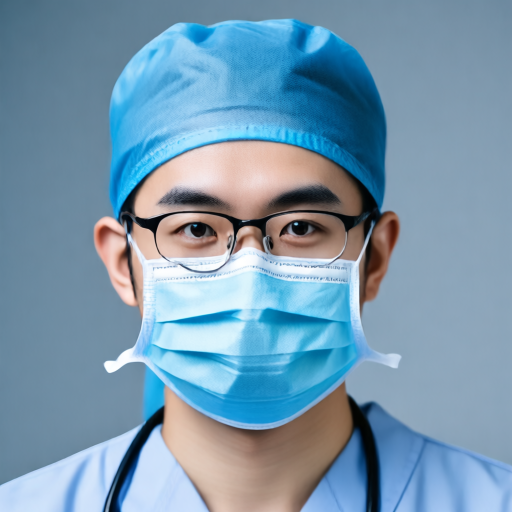}
  \caption{\textit{HUNYUAN}: Representative outputs for each hospital profession without portrait qualifier.}
  \label{fig:hunyuan:genderedImages}
\end{figure}

Figure~\ref{fig:hunyuan:genderedImages} presents representative samples from \textit{HUNYUAN} across the five hospital professions (from left to right: \texttt{cardiologist}, \texttt{hospital director}, \texttt{nurse}, \texttt{paramedic}, and \texttt{surgeon}) generated with the empty portrait qualifier (\texttt{""}). All images were produced at a resolution of 1024\,$\times$\,1024 pixels.

\begin{figure}[!h]
\centering
\includegraphics[width=0.9\textwidth]{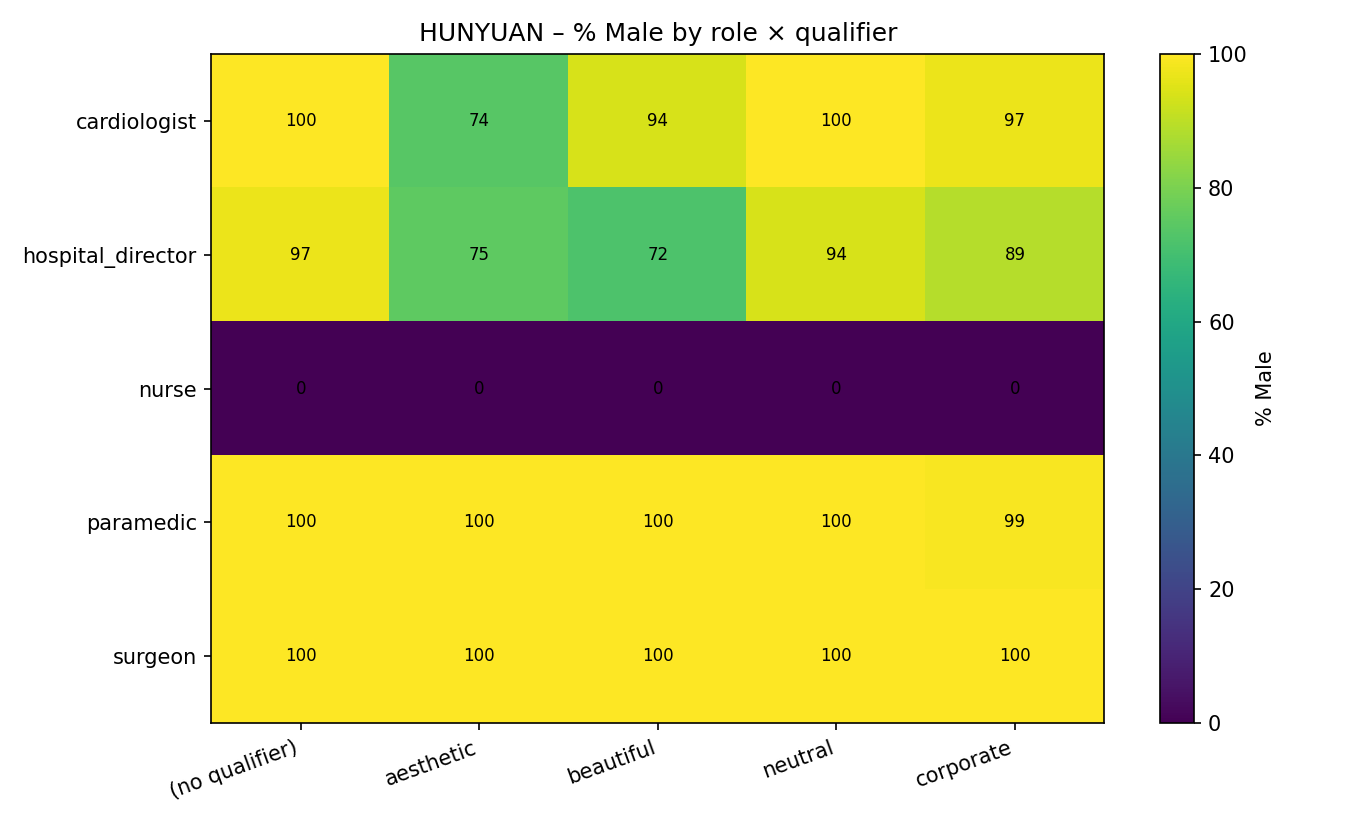}
\caption{\textit{HUNYUAN}: Gender balance across roles and portrait qualifiers.}
\label{fig:HUNYUAN-PS}
\end{figure}

On average (see Figure \ref{fig:HUNYUAN-PS}), \textit{HUNYUAN} yields strongly male-skewed results. Cardiologists are predominantly male (\(93\%\)), hospital directors (\(85\%\)) and paramedics (\(\approx 100\%\)) are almost exclusively male, and surgeons are represented as men in all cases. In sharp contrast, nurses are generated exclusively as women (\(0\%\) male).

The sensitivity analysis (Table~\ref{tab:HUNYUAN:SENSITIVITY}) indicates limited responsiveness to portrait qualifiers. Nurses and surgeons remain fixed (female and male, respectively). Paramedics show virtually no variability (99–100\% male). Moderate variation appears for cardiologists (74–100\% male) and hospital directors (72–97\% male), but male dominance persists across all prompt wordings.

\begin{table}[!h]
\caption{\textit{HUNYUAN}: sensitivity to portrait qualifier.}
\label{tab:HUNYUAN:SENSITIVITY}
\begin{tabular}{lllll}
\toprule
Role & \% Male Average & \% Male Max & \% Male Min & Max–Min \\
\midrule
Cardiologist        & 93 & 100 & 74 & \textbf{26} \\ 
Hospital Director   & 85 & 97 & 72 & 25 \\ 
Nurse               & 0  & 0   & 0  & 0 \\ 
Paramedic           & 100& 100 & 99 & 1 \\ 
Surgeon             & 100& 100 &100 & 0 \\ 
\botrule
\end{tabular}
\end{table}

In summary: (1) \textit{HUNYUAN} enforces rigid stereotypes (female nurses; male surgeons and paramedics). (2) Prompt qualifiers introduce only limited shifts for cardiologists and directors. (3) The model is highly polarized and relatively insensitive to prompt wording.

\subsection{HiDream-I1-dev}

\begin{figure}[!ht]
  \centering
    \includegraphics[width=2.5cm]{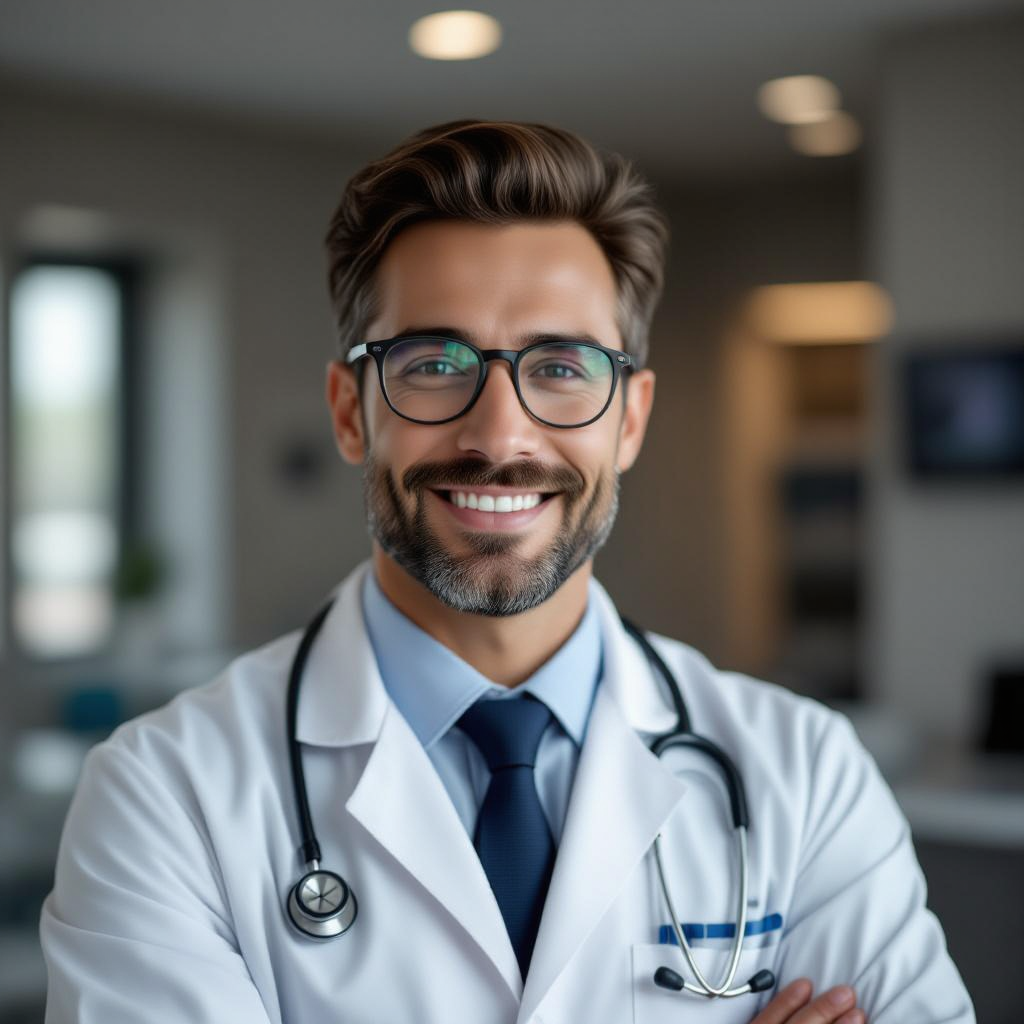}
    \includegraphics[width=2.5cm]{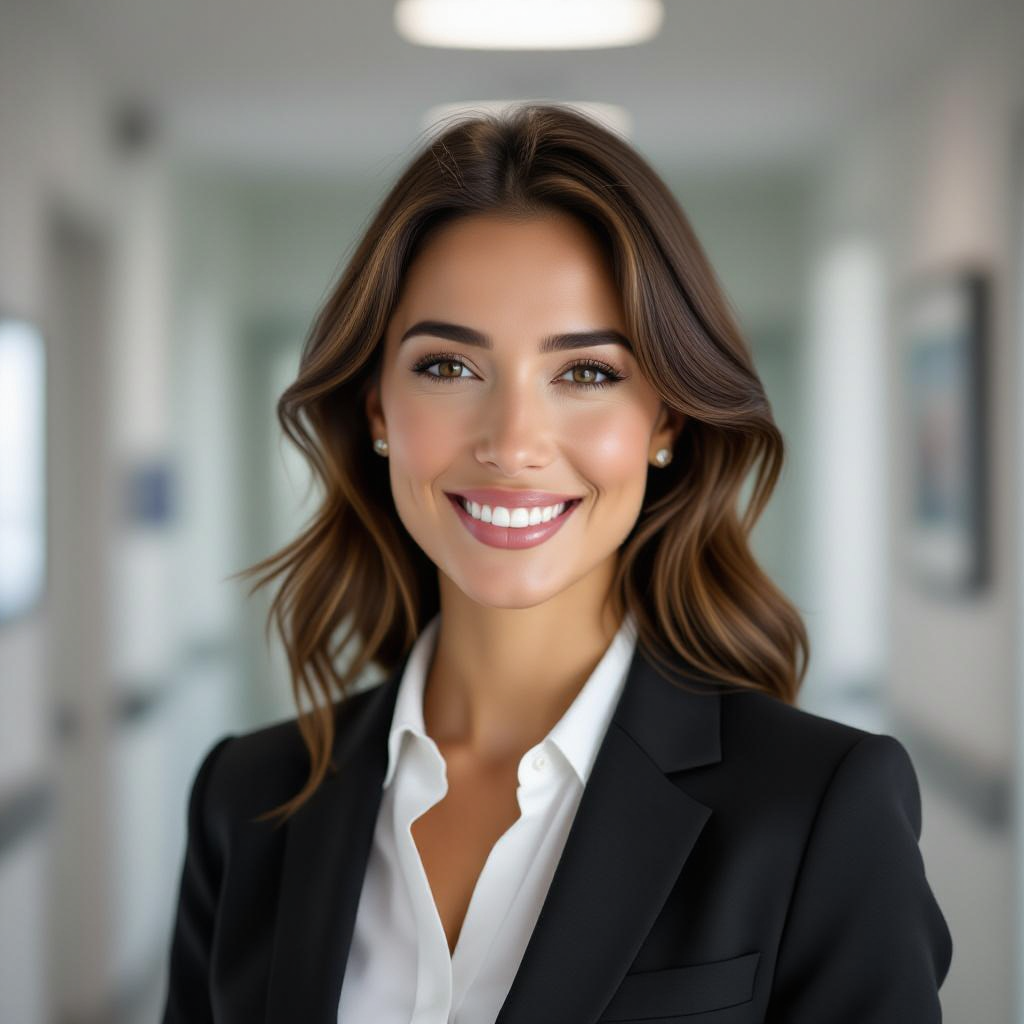}
    \includegraphics[width=2.5cm]{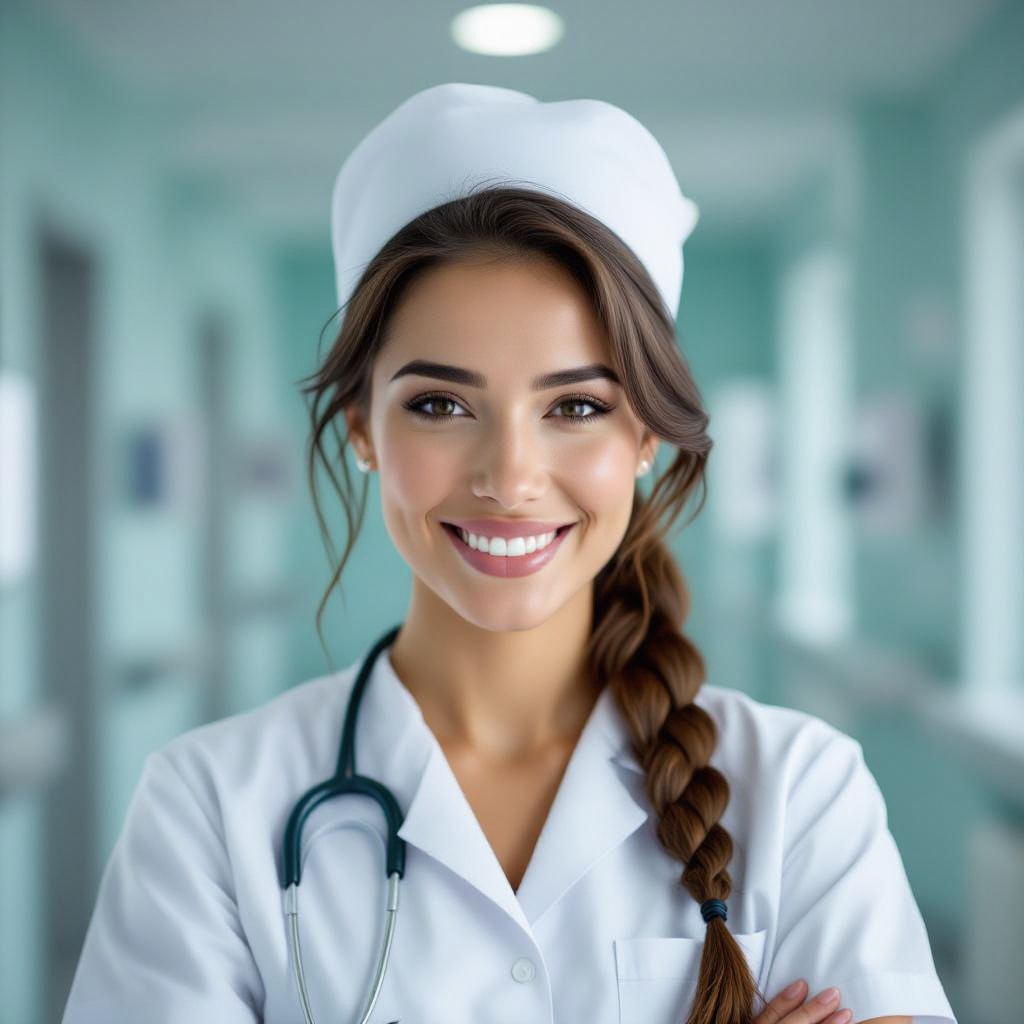}
    \includegraphics[width=2.5cm]{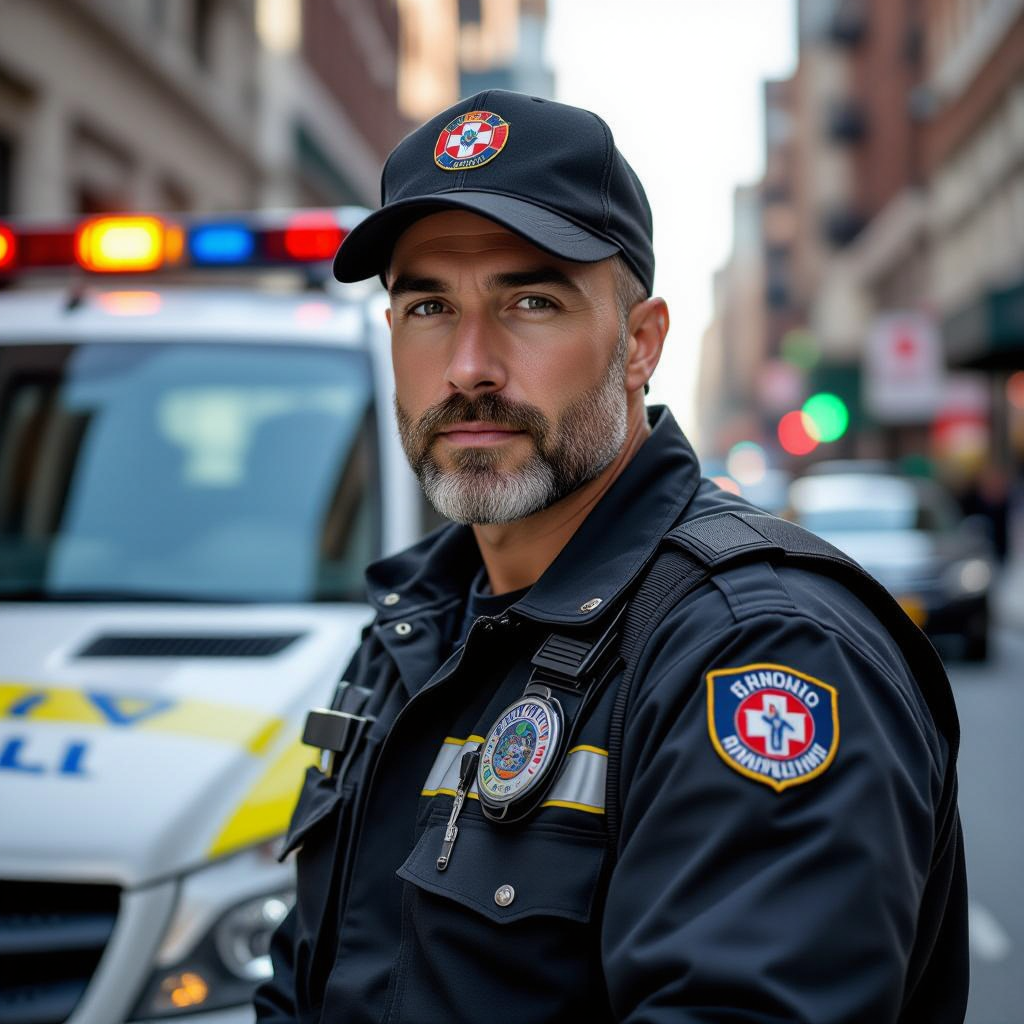}
    \includegraphics[width=2.5cm]{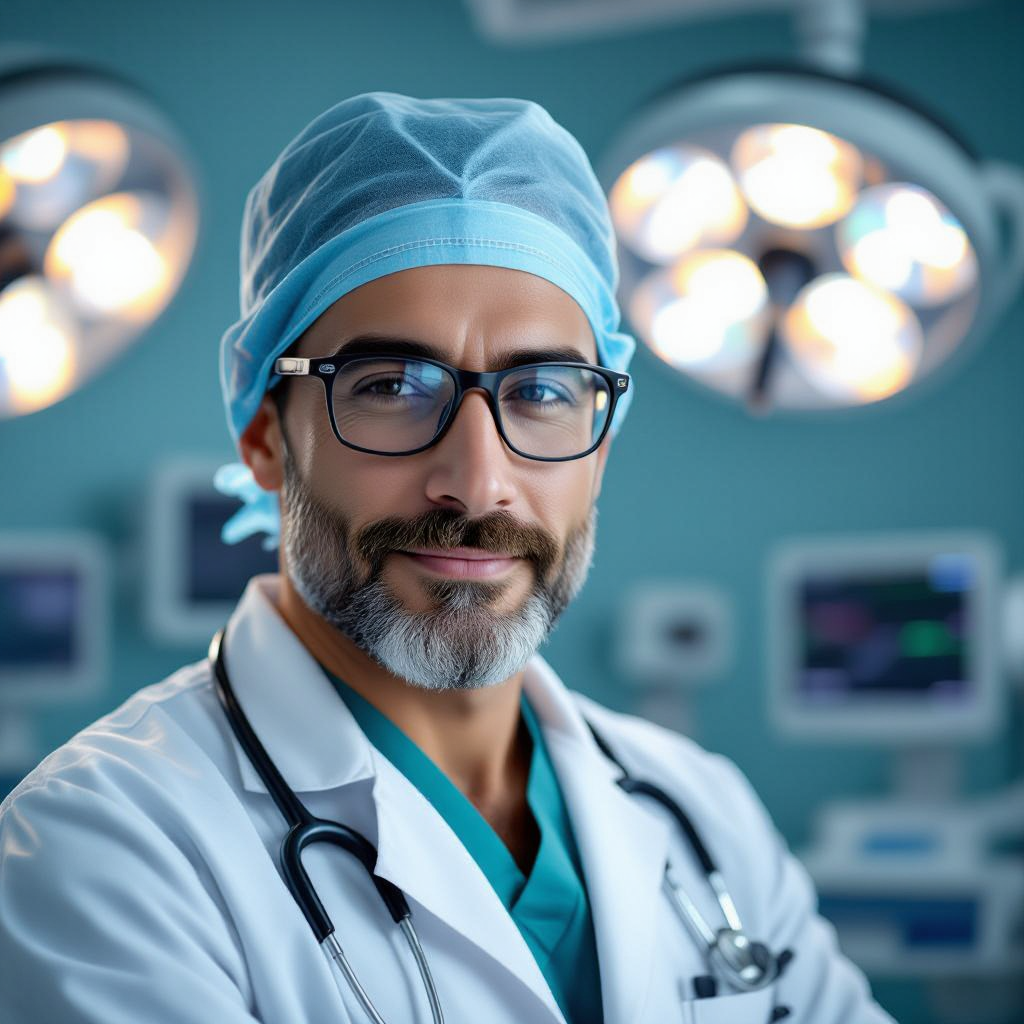}
  \caption{\textit{HIDREAM}: Representative outputs for each hospital profession without portrait qualifier.}
  \label{fig:hidream:genderedImages}
\end{figure}

Figure~\ref{fig:hidream:genderedImages} presents representative samples from \textit{HIDREAM} across the five hospital professions (from left to right: \texttt{cardiologist}, \texttt{hospital director}, \texttt{nurse}, \texttt{paramedic}, and \texttt{surgeon}) generated with the empty portrait qualifier (\texttt{""}). All images were produced at a resolution of 1024\,$\times$\,1024 pixels.

\begin{figure}[h]
\centering
\includegraphics[width=0.9\textwidth]{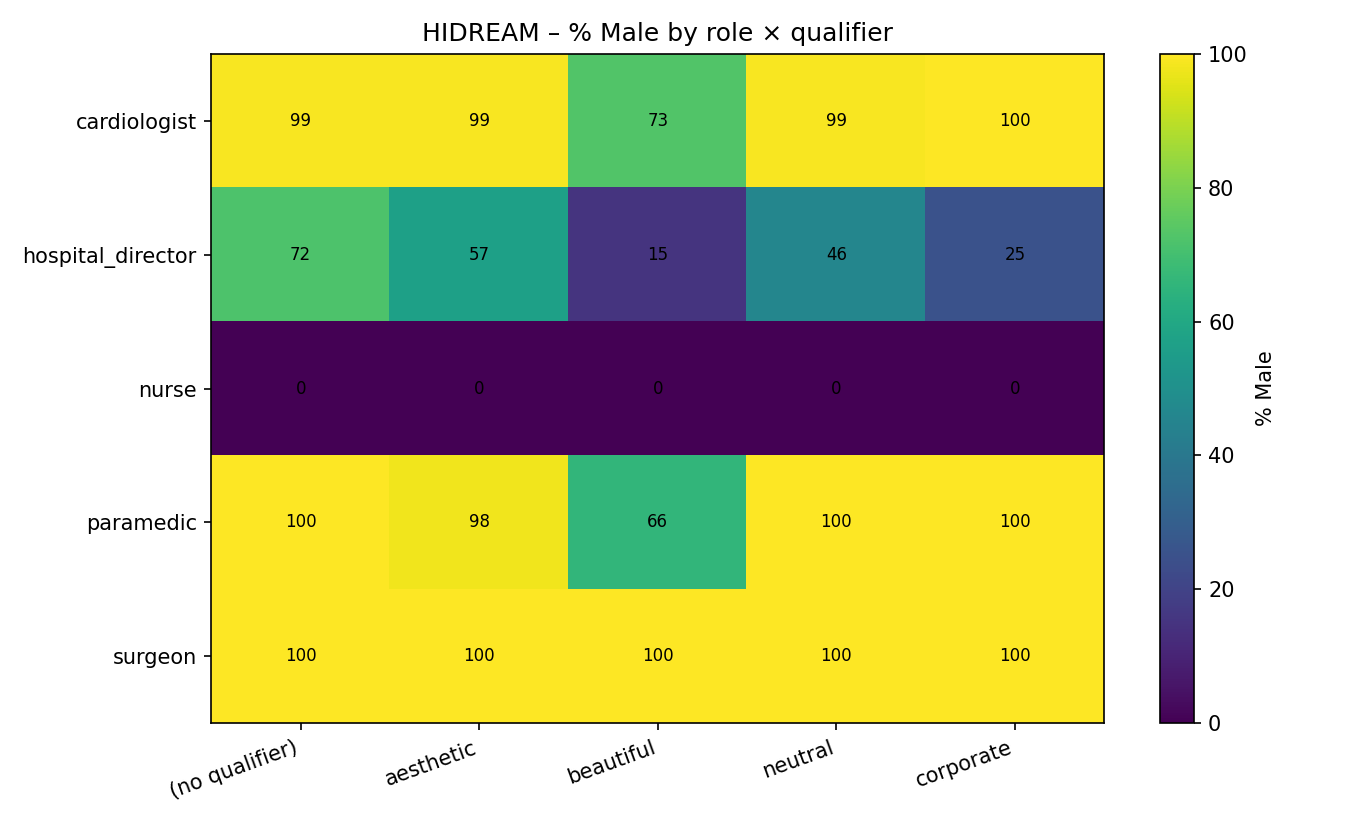}
\caption{\textit{HIDREAM}: Gender balance across roles and portrait qualifiers.}
\label{fig:HIDREAM-PS}
\end{figure}

On average, \textit{HIDREAM} exhibits sharp gender polarizations (see Figure \ref{fig:HIDREAM-PS}). Nurses are always women (\(0\%\) male), and surgeons always men (\(100\%\) male). Cardiologists are overwhelmingly male (\(96\%\)), while hospital directors show partial balance (\(43\%\) male). Paramedics are also male-dominated (\(93\%\)).

The sensitivity analysis (Table~\ref{tab:HIDREAM:SENSITIVITY}) shows uneven responsiveness. Nurses and surgeons remain fixed regardless of qualifier. Cardiologists vary moderately (73–100\% male), while paramedics shift between 66–100\% male. Hospital directors show the strongest effect, from 15\% male (\texttt{beautiful}) to 72\% male (no qualifier), a 57-point spread.

\begin{table}[h]
\caption{\textit{HIDREAM}: sensitivity to portrait qualifier.}
\label{tab:HIDREAM:SENSITIVITY}
\begin{tabular}{lllll}
\toprule
Role & \% Male Average & \% Male Max & \% Male Min & Max–Min \\
\midrule
Cardiologist        & 96 & 100 & 73 & 27 \\ 
Hospital Director   & 43 & 72 & 15 & \textbf{57} \\ 
Nurse               & 0  & 0   & 0  & 0 \\ 
Paramedic           & 93 & 100 & 66 & 34 \\ 
Surgeon             &100 &100  &100 & 0 \\ 
\botrule
\end{tabular}
\end{table}

In summary: (1) \textit{HIDREAM} fixes certain roles to one gender (female nurses, male surgeons). (2) Prompt qualifiers mainly affect hospital directors, and to a lesser extent cardiologists and paramedics. (3) Compared to HUNYUAN, it shows higher variability but equally entrenched stereotypes.

\subsection{Qwen-Image}

\begin{figure}[!ht]
  \centering
    \includegraphics[width=2.5cm]{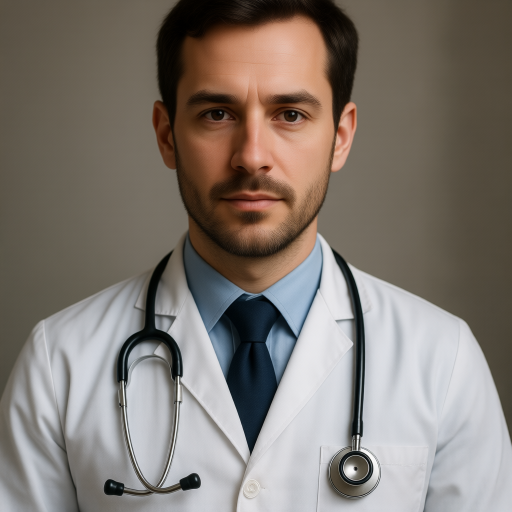}
    \includegraphics[width=2.5cm]{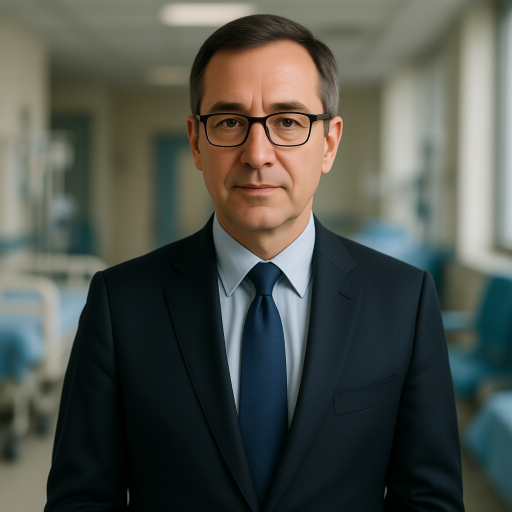}
    \includegraphics[width=2.5cm]{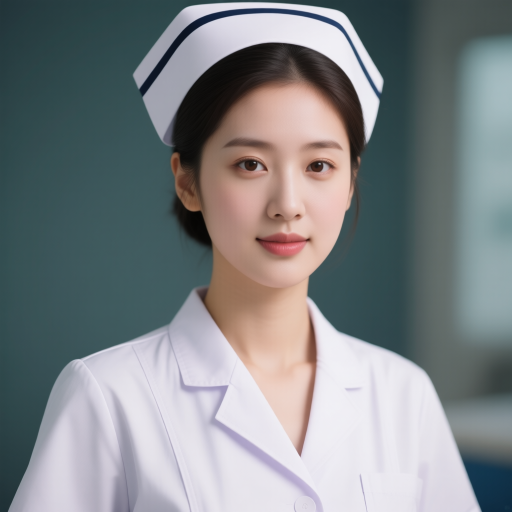}
    \includegraphics[width=2.5cm]{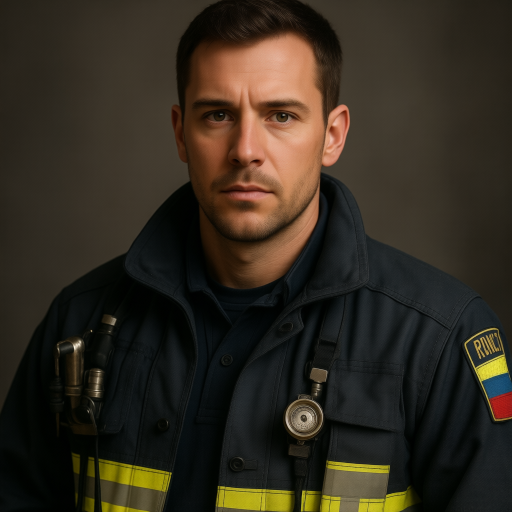}
    \includegraphics[width=2.5cm]{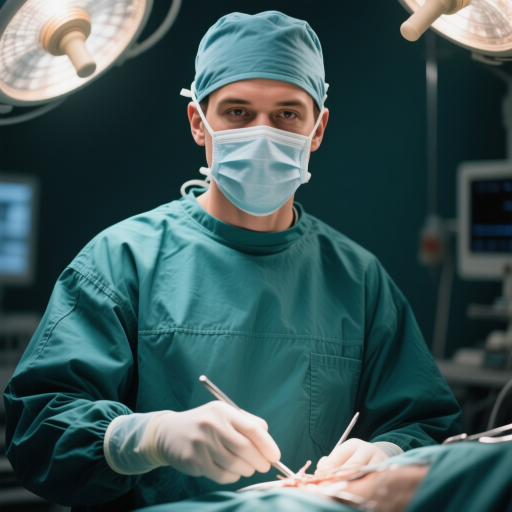}
  \caption{\textit{QWEN}: Representative outputs for each hospital profession without portrait qualifier.}
  \label{fig:qwen:genderedImages}
\end{figure}

Figure~\ref{fig:qwen:genderedImages} presents representative samples from \textit{QWEN} across the five hospital professions (from left to right: \texttt{cardiologist}, \texttt{hospital director}, \texttt{nurse}, \texttt{paramedic}, and \texttt{surgeon}) generated with the empty portrait qualifier (\texttt{""}). All images were produced at a resolution of 1024\,$\times$\,1024 pixels.

\begin{figure}[h]
\centering
\includegraphics[width=0.9\textwidth]{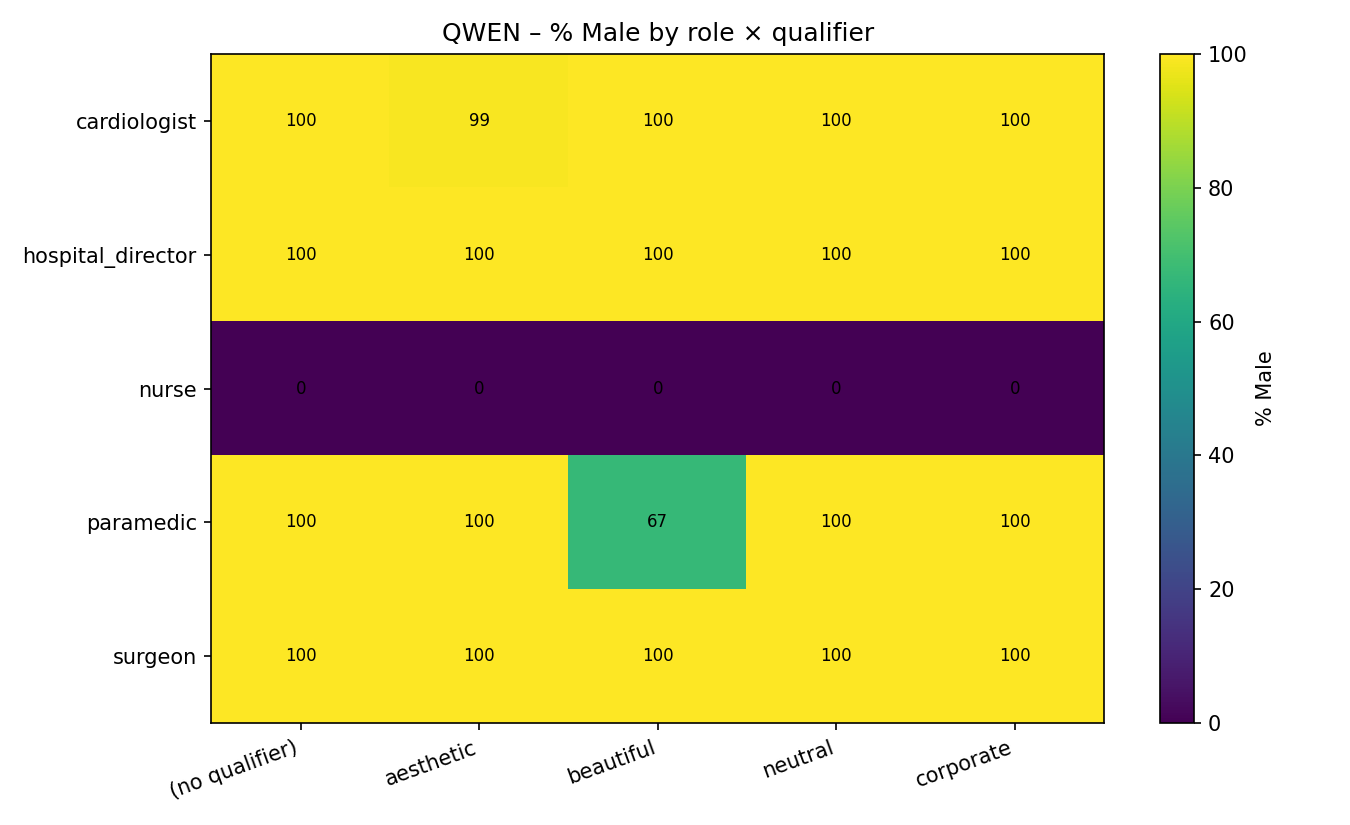}
\caption{\textit{QWEN}: Gender balance across roles and portrait qualifiers.}
\label{fig:QWEN-PS}
\end{figure}

\textit{QWEN} produces the most rigidly polarized outcomes (see Figure \ref{fig:QWEN-PS}). Nurses are always women, and surgeons, directors, and cardiologists are always men (\(\geq 99\%\)). Paramedics are also strongly male (\(93\%\)).

Sensitivity analysis (Table~\ref{tab:QWEN:SENSITIVITY}) confirms near-total rigidity. Nurses, surgeons, and directors show no variation. Cardiologists vary trivially (99–100\% male). Only paramedics show limited responsiveness: 67\% male (\texttt{beautiful}) to 100\% male otherwise.

\begin{table}[h]
\caption{\textit{QWEN}: sensitivity to portrait qualifier.}
\label{tab:QWEN:SENSITIVITY}
\begin{tabular}{lllll}
\toprule
Role & \% Male Average & \% Male Max & \% Male Min & Max–Min \\
\midrule
Cardiologist        & 100 & 100 & 99 & 1 \\ 
Hospital Director   & 100 & 100 &100 & 0 \\ 
Nurse               & 0  & 0   & 0  & 0 \\ 
Paramedic           & 93 & 100 & 67 & \textbf{33} \\ 
Surgeon             &100 &100  &100 & 0 \\ 
\botrule
\end{tabular}
\end{table}

In summary: (1) \textit{QWEN} enforces the strongest binary stereotypes. (2) Prompt qualifiers have virtually no effect. (3) It is the least sensitive and most rigid model among those tested.

\subsection{FLUX.1-dev}

\begin{figure}[!ht]
  \centering
    \includegraphics[width=2.5cm]{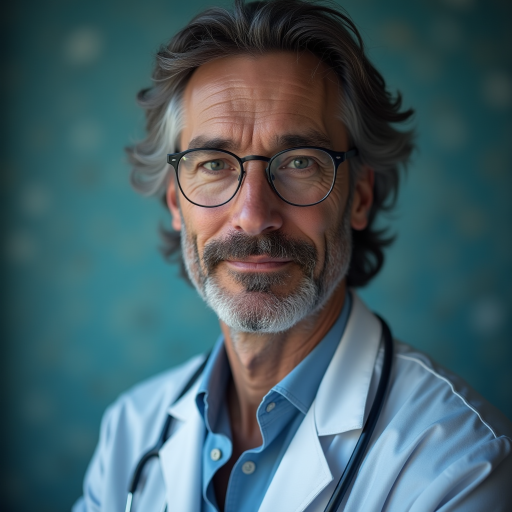}
    \includegraphics[width=2.5cm]{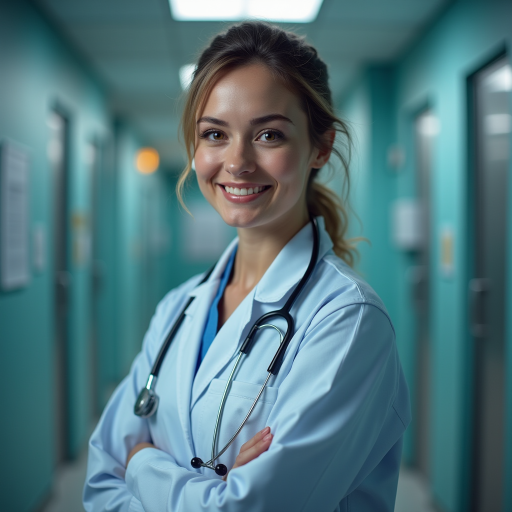}
    \includegraphics[width=2.5cm]{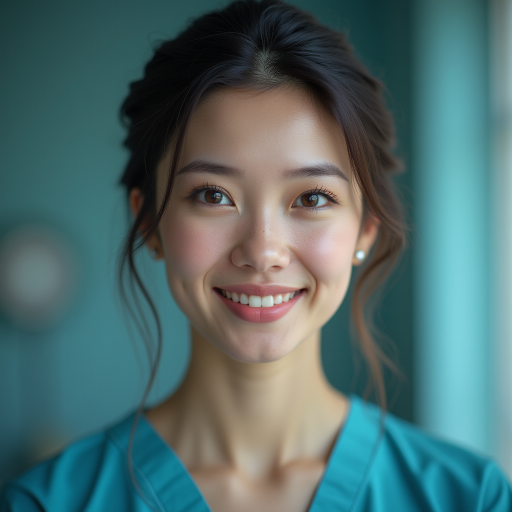}
    \includegraphics[width=2.5cm]{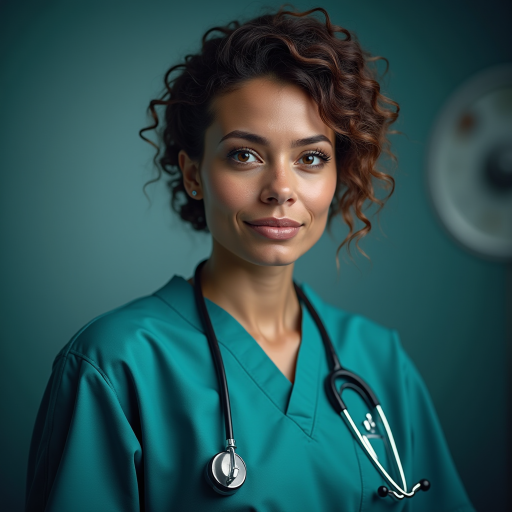}
    \includegraphics[width=2.5cm]{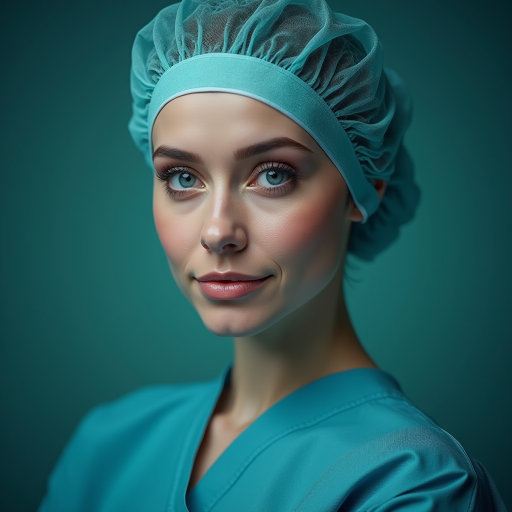}
  \caption{\textit{FLUX}: Representative outputs for each hospital profession without portrait qualifier.}
  \label{fig:flux:genderedImages}
\end{figure}

Figure~\ref{fig:flux:genderedImages} presents representative samples from \textit{FLUX} across the five hospital professions (from left to right: \texttt{cardiologist}, \texttt{hospital director}, \texttt{nurse}, \texttt{paramedic}, and \texttt{surgeon}) generated with the empty portrait qualifier (\texttt{""}). All images were produced at a resolution of 1024\,$\times$\,1024 pixels.

\begin{figure}[!ht]
\centering
\includegraphics[width=0.9\textwidth]{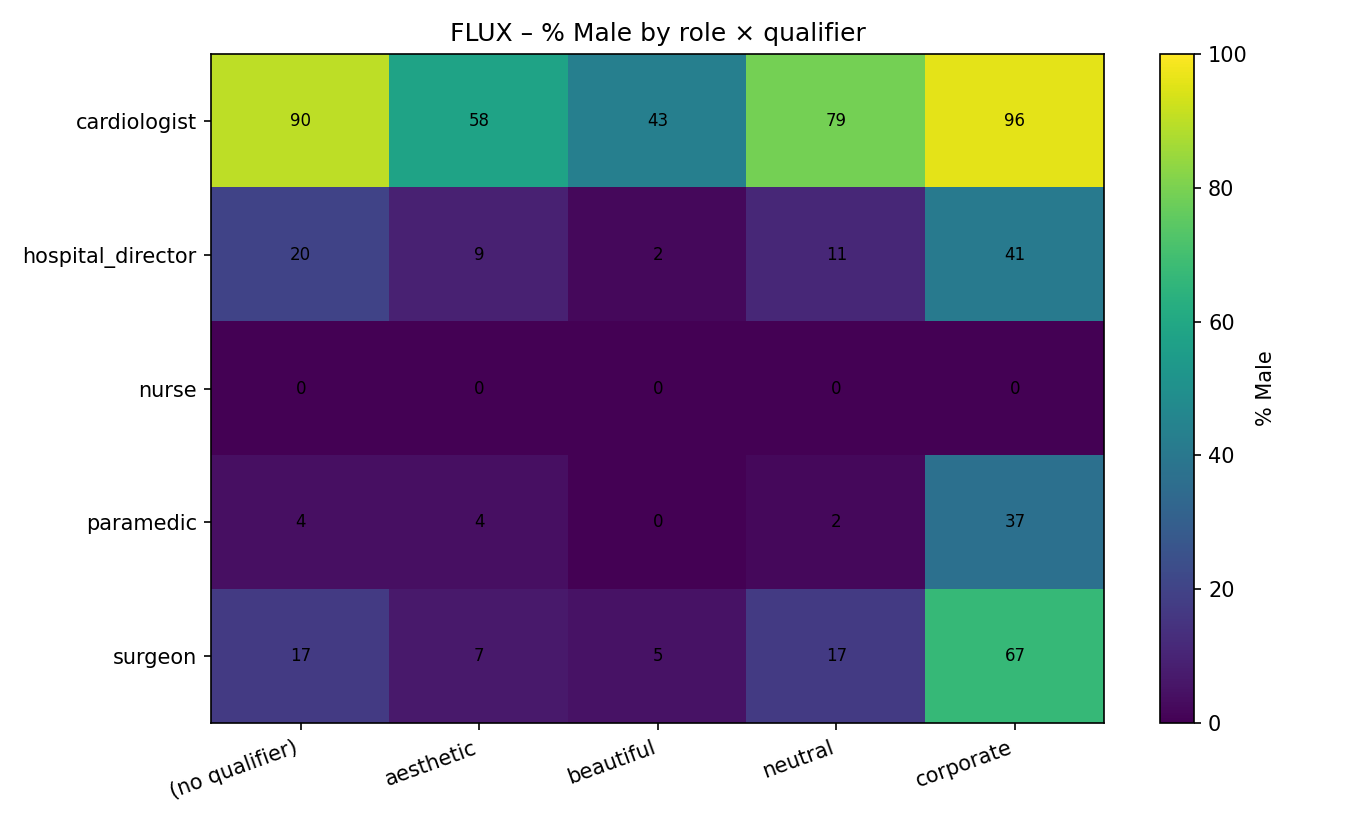}
\caption{\textit{FLUX}: Gender balance across roles and portrait qualifiers.}
\label{fig:FLUX-PS}
\end{figure}

In contrast, \textit{FLUX} generates predominantly female outputs (see Figure \ref{fig:FLUX-PS}). Nurses are always women (\(100\%\)), surgeons are mostly women (\(77\%\)), as are hospital directors (\(83\%\)) and paramedics (\(91\%\)). Only cardiologists skew male (\(73\%\)).

The sensitivity analysis (Table~\ref{tab:FLUX:SENSITIVITY}) reveals high responsiveness. Cardiologists vary widely (43–96\% male), hospital directors from 2–41\% male, and paramedics 0–37\% male. Surgeons show the strongest variation, from 5\% (\texttt{beautiful}) to 67\% male (\texttt{corporate}), a 62-point swing.

\begin{table}[h]
\caption{\textit{FLUX}: sensitivity to portrait qualifier.}
\label{tab:FLUX:SENSITIVITY}
\begin{tabular}{lllll}
\toprule
Role & \% Male Average & \% Male Max & \% Male Min & Max–Min \\
\midrule
Cardiologist        & 73 & 96 & 43 & 53 \\ 
Hospital Director   & 17 & 41 & 2  & 39 \\ 
Nurse               & 0  & 0  & 0  & 0 \\ 
Paramedic           & 9  & 37 & 0  & 37 \\ 
Surgeon             &23  & 67 & 5  & \textbf{62} \\ 
\botrule
\end{tabular}
\end{table}

In summary: (1) \textit{FLUX} is female-skewed in contrast to other models. (2) It is highly sensitive to qualifiers, particularly for surgeons and cardiologists. (3) Compared to QWEN or SDXL, it shows greater flexibility but still reproduces stereotypes.

\subsection{Stable-Diffusion 3.5 Large}

\begin{figure}[!h]
  \centering
    \includegraphics[width=2.5cm]{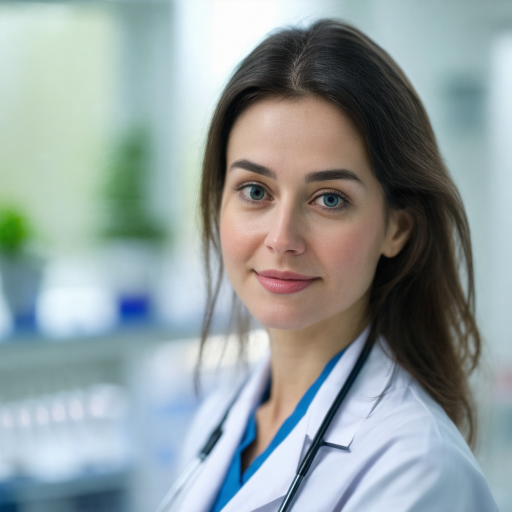}
    \includegraphics[width=2.5cm]{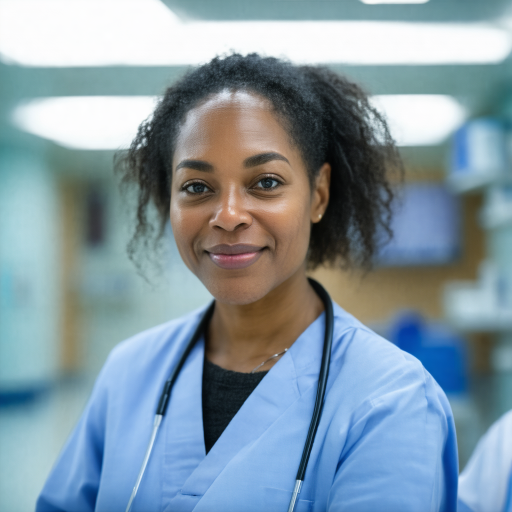}.
    \includegraphics[width=2.5cm]{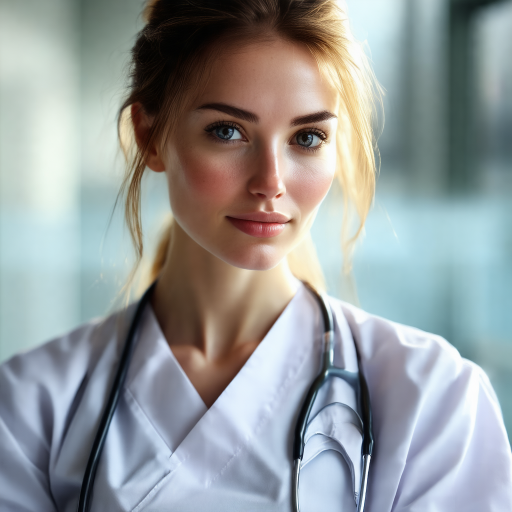}
    \includegraphics[width=2.5cm]{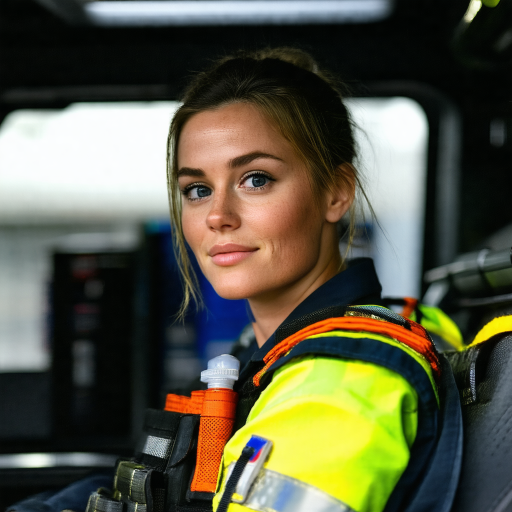}
    \includegraphics[width=2.5cm]{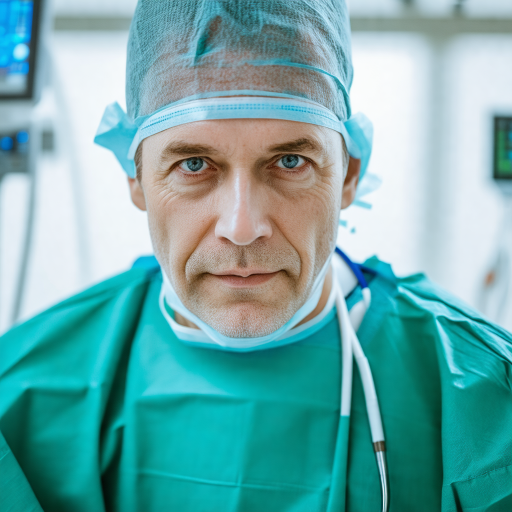}
  \caption{\textit{SD3.5}: Representative outputs for each hospital profession without portrait qualifier.}
  \label{fig:SD3.5:genderedImages}
\end{figure}

Figure~\ref{fig:SD3.5:genderedImages} presents representative samples from \textit{SD3.5} across the five hospital professions (from left to right: \texttt{cardiologist}, \texttt{hospital director}, \texttt{nurse}, \texttt{paramedic}, and \texttt{surgeon}) generated with the empty portrait qualifier (\texttt{""}). All images were produced at a resolution of 1024\,$\times$\,1024 pixels.

\begin{figure}[!h]
\centering
\includegraphics[width=0.9\textwidth]{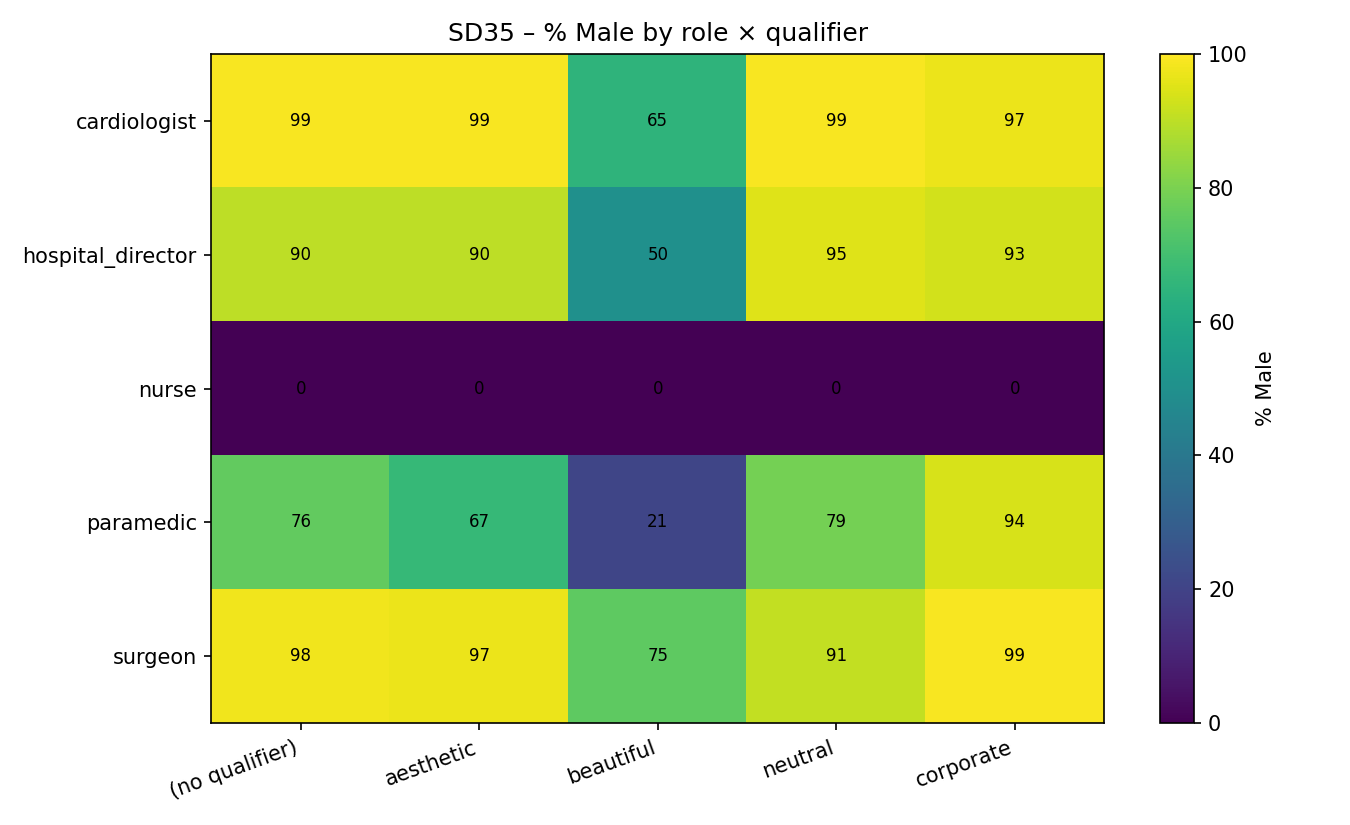}
\caption{\textit{SD3.5}: Gender balance across roles and portrait qualifiers.}
\label{fig:SD3.5-PS}
\end{figure}

\textit{SD3.5} produces strongly male-dominated results (see Figure \ref{fig:SD3.5-PS}). Cardiologists (\(92\%\)), surgeons (\(92\%\)), and hospital directors (\(85\%\)) are overwhelmingly male. Nurses are exclusively female, while paramedics are somewhat more balanced (\(67\%\) male on average).

The sensitivity analysis (Table~\ref{tab:SD3.5:SENSITIVITY}) highlights strong qualifier effects for paramedics (21–94\% male, 73-point swing) and moderate effects for directors (50–90\% male, 49-point swing). Surgeons (75–99\%) and cardiologists (65–99\%) vary less but remain male-dominated. Nurses stay fixed at \(0\%\) male.

\begin{table}[h]
\caption{\textit{SD3.5}: sensitivity to portrait qualifier.}
\label{tab:SD3.5:SENSITIVITY}
\begin{tabular}{lllll}
\toprule
Role & \% Male Average & \% Male Max & \% Male Min & Max–Min \\
\midrule
Cardiologist        & 92 & 99 & 65 & 34 \\ 
Hospital Director   & 85 & 90 & 50 & 49 \\ 
Nurse               & 0  & 0  & 0  & 0 \\ 
Paramedic           & 67 & 94 & 21 & \textbf{73} \\ 
Surgeon             & 92 & 99 & 75 & 24 \\ 
\botrule
\end{tabular}
\end{table}

In summary: (1) \textit{SD3.5} enforces strong male stereotypes for most roles. (2) Paramedics are the most sensitive to qualifiers, shifting from female- to male-dominated. (3) Compared to QWEN, it allows more variability but still amplifies occupational stereotypes.

\subsection{Stable-Diffusion-XL}

\begin{figure}[!h]
  \centering
    \includegraphics[width=2.5cm]{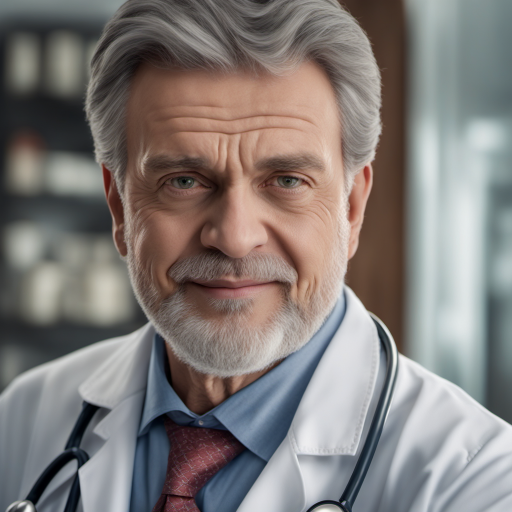}
    \includegraphics[width=2.5cm]{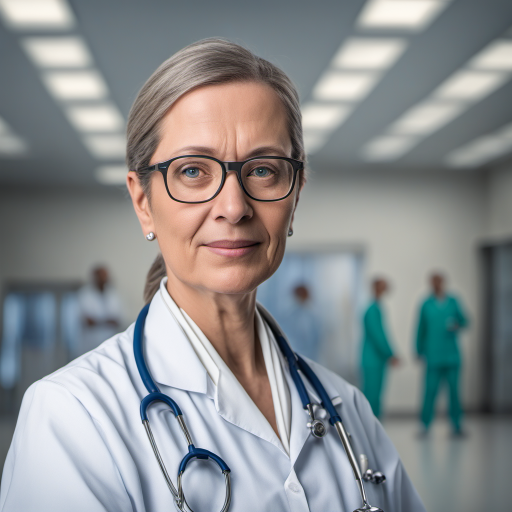}
    \includegraphics[width=2.5cm]{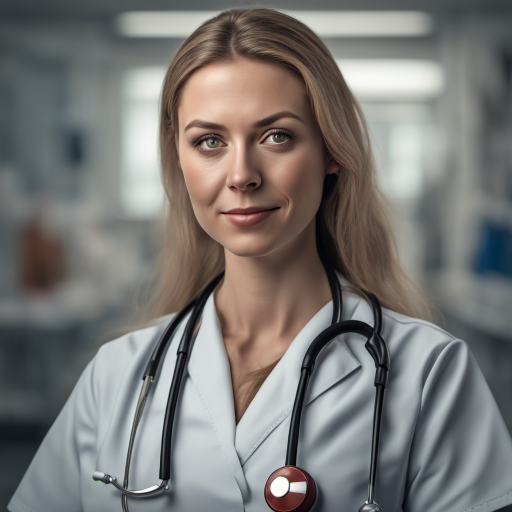}
    \includegraphics[width=2.5cm]{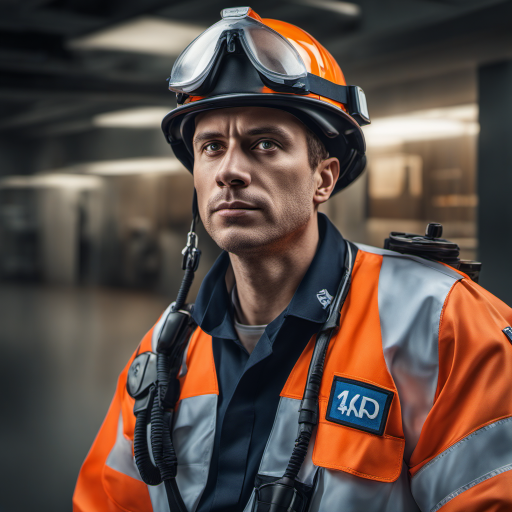}
    \includegraphics[width=2.5cm]{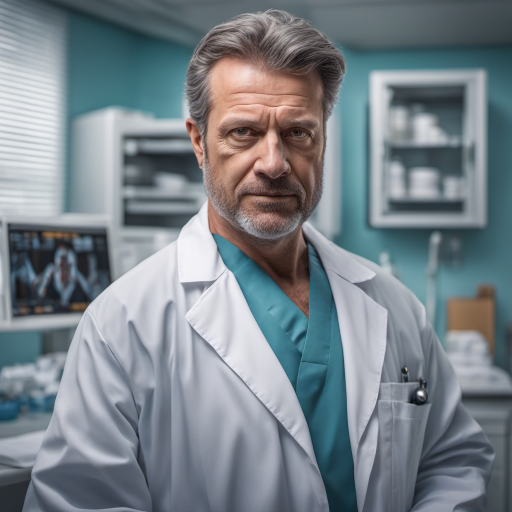}
  \caption{\textit{SDXL}: Representative outputs for each hospital profession without portrait qualifier.}
  \label{fig:sdxl:genderedImages}
\end{figure}

Figure~\ref{fig:sdxl:genderedImages} presents representative samples from \textit{SDXL} across the five hospital professions (from left to right: \texttt{cardiologist}, \texttt{hospital director}, \texttt{nurse}, \texttt{paramedic}, and \texttt{surgeon}) generated with the empty portrait qualifier (\texttt{""}). All images were produced at a resolution of 1024\,$\times$\,1024 pixels.

\begin{figure}[!h]
\centering
\includegraphics[width=0.9\textwidth]{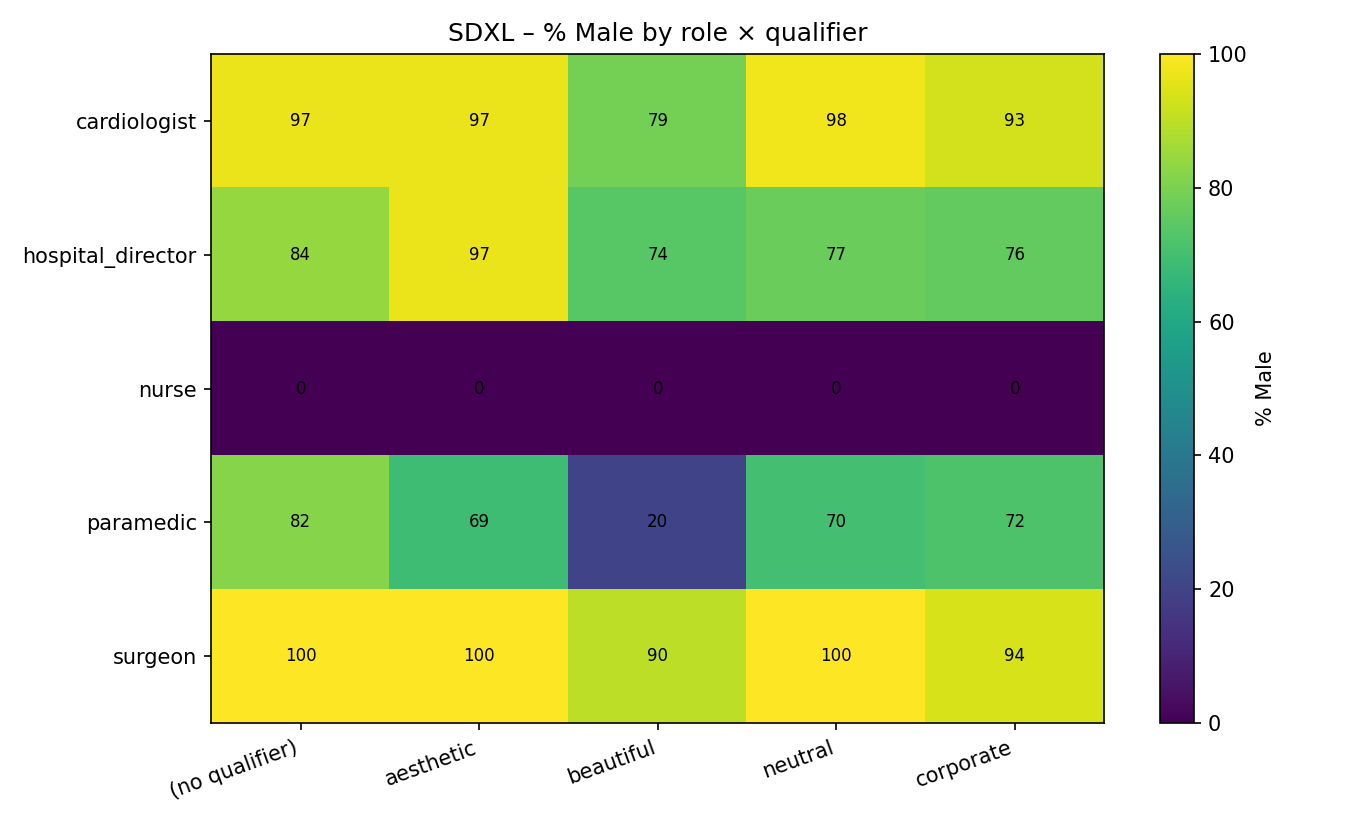}
\caption{\textit{SDXL}: Gender balance across roles and portrait qualifiers.}
\label{fig:SDXL-PS}
\end{figure}

Like SD3.5, \textit{SDXL} generates strongly male-skewed results (see Figure \ref{fig:SDXL-PS}). Cardiologists average \(93\%\) male, surgeons \(97\%\), and hospital directors \(82\%\). Nurses are consistently female, while paramedics show more balance (\(63\%\) male on average).

The sensitivity analysis (Table~\ref{tab:SDXL:SENSITIVITY}) shows minor effects for cardiologists (79–98\%) and surgeons (90–100\%), moderate effects for directors (74–97\%), and strong effects for paramedics (20–82\%, 62-point swing). Nurses remain fixed.

\begin{table}[h]
\caption{\textit{SDXL}: sensitivity to portrait qualifier.}
\label{tab:SDXL:SENSITIVITY}
\begin{tabular}{lllll}
\toprule
Role & \% Male Average & \% Male Max & \% Male Min & Max–Min \\
\midrule
Cardiologist        & 93 & 98 & 79 & 19 \\ 
Hospital Director   & 82 & 97 & 74 & 23 \\ 
Nurse               & 0  & 0  & 0  & 0 \\ 
Paramedic           & 63 & 82 & 20 & \textbf{62} \\ 
Surgeon             & 97 &100 & 90 & 10 \\ 
\botrule
\end{tabular}
\end{table}

In summary: (1) \textit{SDXL} systematically reinforces male-coded roles and female-coded nurses. (2) Paramedics show the highest sensitivity, shifting from female- to male-dominated. (3) Compared to SD3.5, the magnitude of variability is slightly reduced but the overall bias remains.

\subsection{Comparative analysis across models}
\label{sec:comparative}

A cross-model comparison highlights both convergences and divergences in gender distribution patterns. 
Table~\ref{tab:COMPARATIVE:ROLES} summarizes the average proportion of male images generated by each model across the five professional roles. Clear tendencies emerge: 
\begin{itemize}
    \item \textit{QWEN}, \textit{SD3.5}, and \textit{SDXL} strongly favor male representations for most roles (cardiologist, surgeon, hospital director), while nurses are exclusively depicted as female. 
    \item \textit{HIDREAM} and \textit{HUNYUAN} also produce male-dominated distributions but with some role-specific variation.
    \item In contrast, \textit{FLUX} is the only model to generate a female majority overall, though with role-specific exceptions (notably male-skewed cardiologists versus female-skewed nurses, paramedics, and directors).
\end{itemize}

\begin{table}[h]
\caption{Average proportion of male images across roles for each model (\%).}
\label{tab:COMPARATIVE:ROLES}
\begin{tabular*}{\textwidth}{@{\extracolsep\fill}lcccccc}
\toprule
Role & FLUX & HIDREAM & QWEN & SD3.5 & SDXL & HUNYUAN \\
\midrule
Cardiologist        & 73 & 96 & 100 & 92 & 93 & 93 \\
Hospital Director   & 17 & 43 & 100 & 85 & 82 & 85 \\
Paramedic           &  9 & 93 &  93 & 67 & 63 & 100 \\
Surgeon             & 23 &100 & 100 & 92 & 97 & 100 \\
Nurse               &  0 &  0 &   0 &  0 &  0 &  0 \\
\botrule
\end{tabular*}
\end{table}

\begin{figure}[h]
\centering
\includegraphics[width=\linewidth]{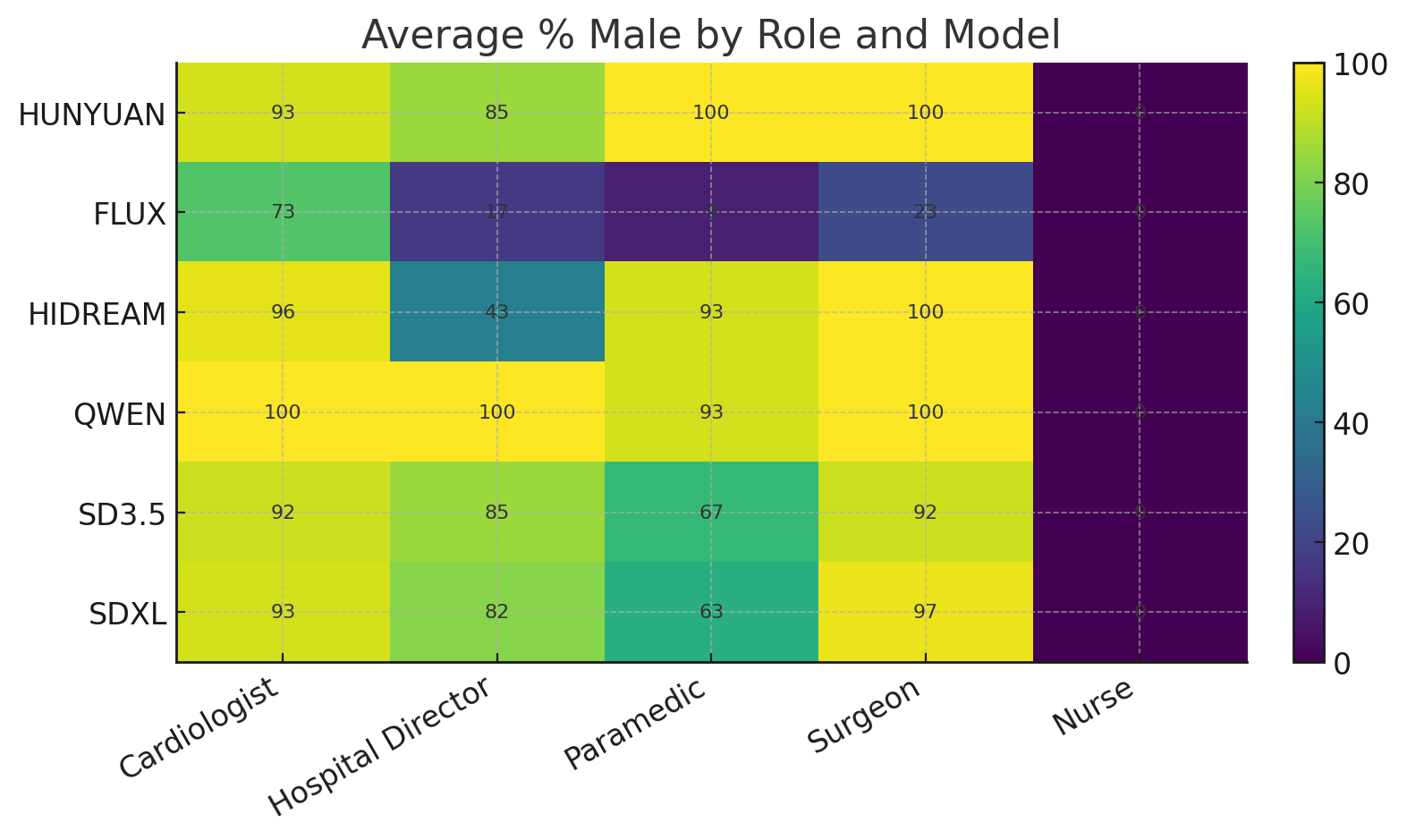}
\caption{Average proportion of male images by role across models (HUNYUAN, FLUX, HIDREAM, QWEN, SD3.5, SDXL). Higher values indicate stronger male dominance.}
\label{fig:GLOBAL-AVG}
\end{figure}

Sensitivity to portrait qualifiers also differs sharply across models (Table~\ref{tab:COMPARATIVE:SENSITIVITY}). \textit{QWEN} and \textit{HUNYUAN} show almost no responsiveness to prompt variation, with gender outcomes remaining fixed for most roles. \textit{HIDREAM} displays role-specific sensitivity, particularly for hospital directors (57-point variation). \textit{SD3.5} shows the strongest qualifier effect for paramedics (73 points), while \textit{SDXL} exhibits a similar pattern with 62 points of variation. \textit{FLUX} demonstrates high sensitivity for both surgeons (62 points) and cardiologists (53 points). 

\begin{table}[h]
\caption{Maximum variation in gender balance across portrait qualifiers (percentage points).}
\label{tab:COMPARATIVE:SENSITIVITY}
\begin{tabular*}{\textwidth}{@{\extracolsep\fill}lcccccc}
\toprule
Role & FLUX & HIDREAM & QWEN & SD3.5 & SDXL & HUNYUAN \\
\midrule
Cardiologist        & 53 & 27 &  1 & 34 & 19 & 26 \\
Hospital Director   & 39 & 57 &  0 & 49 & 23 & 25 \\
Paramedic           & 37 & 34 & 33 & \textbf{73} & \textbf{62} & 1 \\
Surgeon             & \textbf{62} &  0 &  0 & 24 & 10 & 0 \\
Nurse               &  0 &  0 &  0 &  0 &  0 & 0 \\
\botrule
\end{tabular*}
\end{table}

\begin{figure}[h]
\centering
\includegraphics[width=\linewidth]{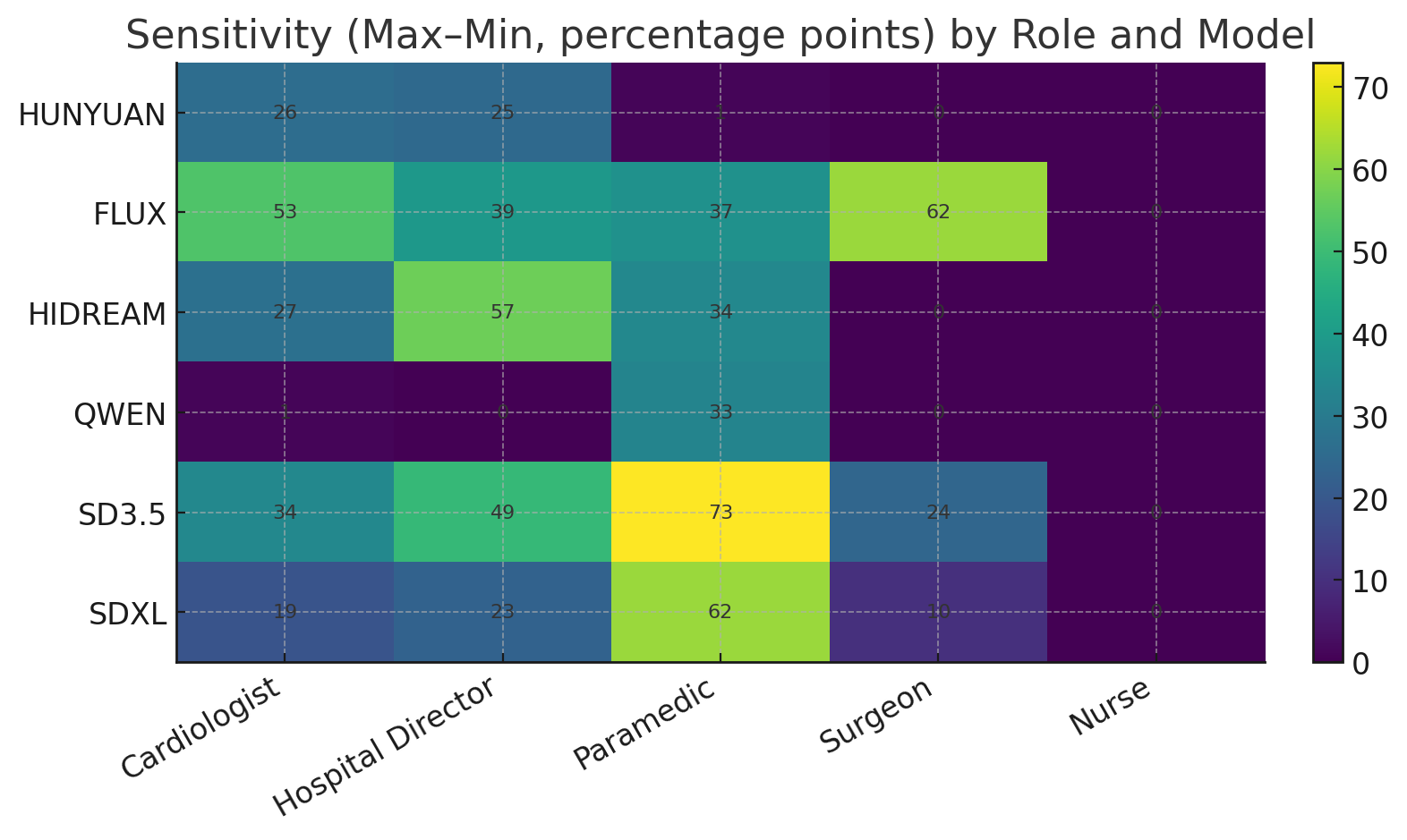}
\caption{Sensitivity of gender balance to portrait qualifiers (Max--Min, percentage points) by role across models. Higher values indicate stronger dependence on prompt wording.}
\label{fig:GLOBAL-SENS}
\end{figure}

In summary, the comparative analysis reveals three main findings:
\begin{enumerate}
    \item All models reproduce strong occupational gender stereotypes, with nurses consistently female and surgeons overwhelmingly male (except in \textit{FLUX}, which leans female overall).
    \item The influence of portrait qualifiers varies widely: minimal in \textit{QWEN} and \textit{HUNYUAN}, moderate and role-specific in \textit{HIDREAM}, and substantial in both \textit{FLUX}, \textit{SD3.5}, and \textit{SDXL}.
    \item These patterns suggest that while bias is systematic across models, its expression differs: some models enforce rigid stereotypes independent of prompt formulation (\textit{QWEN}, \textit{HUNYUAN}), while others amplify or mitigate biases depending on prompt wording (\textit{FLUX}, \textit{SD3.5}, \textit{SDXL}).
\end{enumerate}

\section{Discussion}
\label{sec:discussion}

\subsection{Gender Distribution Across Professions: Evidence of Bias}

Our comparative analysis across six state-of-the-art text-to-image models (\textit{HUNYUAN}, \textit{FLUX}, \textit{HIDREAM}, \textit{QWEN}, \textit{SD3.5}, \textit{SDXL}) reveals systematic occupational gender stereotypes (see Section~\ref{sec:results}). Figure~\ref{fig:GLOBAL-AVG} and Table~\ref{tab:COMPARATIVE:ROLES} summarize these distributions. Clear and consistent patterns emerge:

\begin{itemize}
    \item \textbf{Nurses}: All models without exception generated $100\%$ female images. While this aligns with traditional gender expectations, it completely erases the real but minority presence of male nurses worldwide. 
    \item \textbf{Surgeons}: With the exception of \textit{FLUX}, which leaned female overall, all models produced surgeons almost exclusively as men (92--100\%). This outcome exaggerates the underrepresentation of women observed in real-world statistics (e.g., 17\% female surgeons in the UK \cite{RoyalCollegeSurgeons2022}).
    \item \textbf{Cardiologists and Hospital Directors}: \textit{QWEN}, \textit{SD3.5}, \textit{SDXL}, and \textit{HUNYUAN} generate nearly all cardiologists and directors as male. \textit{HIDREAM} shows partial variation, while \textit{FLUX} reverses the trend with female-skewed outputs (83\% of hospital directors female). 
    \item \textbf{Paramedics}: Results vary considerably, with male dominance in \textit{HIDREAM}, \textit{QWEN}, \textit{SD3.5}, \textit{SDXL}, and \textit{HUNYUAN}, versus a female majority in \textit{FLUX}.
\end{itemize}

These results highlight two types of biases: rigid stereotypes (e.g., female nurses, male surgeons) and model-specific skews (e.g., \textit{FLUX} producing more female figures than other models). Both patterns deviate substantially from real-world distributions, sometimes exaggerating existing inequalities.

\subsection{The Role of the Portrait Qualifier}

In addition to profession-driven differences, portrait qualifiers (\texttt{""}, \texttt{corporate}, \texttt{neutral}, \texttt{aesthetic}, \texttt{beautiful}) exert a strong influence on gender representation. This effect is summarized in Figure~\ref{fig:GLOBAL-SENS} and Table~\ref{tab:COMPARATIVE:SENSITIVITY}. 

Two clear trends emerge:

\begin{itemize}
    \item \textbf{High sensitivity}: In \textit{FLUX}, \textit{SD3.5}, and \textit{SDXL}, certain professions exhibit large variations depending on the qualifier. For instance, male representations among surgeons in \textit{FLUX} range from $5\%$ (“beautiful”) to $67\%$ (“corporate”), and paramedics in \textit{SD3.5} shift from $21\%$ (“beautiful”) to $94\%$ (“corporate”). 
    \item \textbf{Low sensitivity}: \textit{QWEN} and \textit{HUNYUAN} are almost insensitive to qualifiers, with gender outcomes fixed regardless of wording. \textit{HIDREAM} lies in between, with strong variation for hospital directors (57 points) but rigidity in other roles such as surgeons and nurses.
\end{itemize}

This pattern reveals another layer of implicit bias: terms like \texttt{corporate} tend to be semantically linked with male representations, whereas \texttt{beautiful} skews strongly female. Beyond professional labels, descriptive framing thus profoundly influences how gender is expressed by generative models, potentially reinforcing stereotypical associations between gender, aesthetics, and authority.

\subsection{Ethical Implications}

The outputs of these models mirror---and in some cases exaggerate---gender stereotypes, embedding them into visual archetypes. Such biases do not merely reflect existing social inequalities but actively shape perceptions of professional identity. In healthcare, where gender equity remains a critical issue, the uncritical use of these images could reinforce stereotypes about competence, leadership, or caregiving roles.

If adopted in communication materials, educational resources, or recruitment tools, such biased outputs risk legitimizing occupational imbalances. This concern extends to other professional fields tested in our broader analysis (e.g., law, teaching, sports, arts), suggesting that the problem is model-agnostic.

Given this evidence, we identify a pressing need for bias mitigation strategies in generative AI: 
\begin{enumerate}
    \item \textbf{Balanced generation defaults}: Offering options to generate balanced outputs (e.g., 50/50 male–female) in professional contexts. 
    \item \textbf{Prompt suggestion systems}: Encouraging users to adopt neutral or diversity-promoting prompt formulations. 
    \item \textbf{User agency}: Explicitly specifying gender or ethnicity in prompts can counteract default biases, though this requires awareness and responsibility from end-users. 
\end{enumerate}

Overall, while text-to-image models open new possibilities for content creation, their current outputs risk reproducing and amplifying harmful stereotypes. Addressing these issues requires both technical solutions at the model-design level and critical awareness on the part of practitioners and end-users.

\section{Conclusion}
\label{sec:conclusion}

This study provides a systematic evaluation of gender representation in six state-of-the-art text-to-image models: \textit{HunyuanImage 2.1}, \textit{FLUX.1-dev}, \textit{HiDream-I1-dev}, \textit{Qwen-Image}, \textit{Stable-Diffusion 3.5 Large}, and \textit{Stable-Diffusion-XL}. By examining hospital-related professions across multiple prompt configurations, we demonstrate that all models reproduce strong occupational stereotypes, with consistent patterns such as exclusively female nurses and overwhelmingly male surgeons. Nevertheless, important differences emerge across models: \textit{Qwen-Image}, \textit{SD3.5}, and \textit{SDXL} enforce particularly rigid male dominance; \textit{HunyuanImage 2.1} and \textit{HiDream-I1-dev} display intermediate but still highly skewed outcomes; while \textit{FLUX.1-dev} tends toward female-skewed outputs. These differences illustrate that bias is not uniform but model-specific, reflecting architectural choices, training corpora, and sensitivity to prompt formulation.  

Beyond professional labels, our analysis shows that portrait qualifiers (\texttt{corporate}, \texttt{beautiful}, etc.) exert a significant influence on gender balance. Terms associated with authority or professionalism tend to reinforce male representations, whereas aesthetic qualifiers skew strongly female. This reveals a deeper semantic layer of bias in generative AI: linguistic framing shapes not only visual style but also demographic outcomes.  

Taken together, these findings raise important concerns for fairness, accountability, and transparency in generative AI. In professional contexts such as healthcare, biased image outputs risk reinforcing stereotypes and undermining equity efforts. To mitigate such risks, we argue for the development of bias-aware generation strategies, including balanced default settings, prompt guidance systems, and mechanisms that allow users to specify demographic attributes in a transparent and ethical way.  

Future work should expand beyond gender to examine intersectional dimensions such as ethnicity, age, or disability, and should systematically compare open-weight and proprietary models. Additionally, the impact of such biases on downstream applications---from education to recruitment---deserves closer investigation. Addressing these issues is essential to ensure that generative AI contributes not to the perpetuation of stereotypes, but to the creation of more diverse, inclusive, and equitable representations.  

\begin{appendices}

\section{Results Tables}\label{secA1}
This appendix provides the full, unaggregated results for each model and each prompt configuration. For every combination of \emph{model} (\textit{HunyuanImage 2.1}: Table~\ref{tab:HUNYUAN}, \textit{HiDream-I1-dev}: Table~\ref{tab:HIDREAM}, \textit{Qwen-Image}: Table~\ref{tab:QWEN},\textit{FLUX.1-dev}: Table~\ref{tab:FLUX},  \textit{Stable-Diffusion-XL}: Table~\ref{tab:SDXL}), \emph{hospital role} (cardiologist, hospital director, nurse, paramedic, surgeon), and \emph{portrait qualifier} (\texttt{""}, \texttt{aesthetic}, \texttt{beautiful}, \texttt{corporate}, \texttt{neutral}), we report the raw counts of male and female images together with the total number of generated images.

\begin{table}[!ht]
\caption{HunyuanImage 2.1: Gender balance of generated images by hospital role and portrait qualifier}\label{tab:HUNYUAN}
\begin{tabular*}{\textwidth}{@{\extracolsep\fill}llllll}
\toprule%
 Model & Qualifier & Role & Male Images & Female Images & Total Images \\
\midrule
  HUNYUAN & aesthetic   & cardiologist        & M= 74  & F=  26  & T= 100    \\
  &                     & hospital director   & M= 75  & F= 25  & T= 100    \\
  &                     & paramedic           & M= 100  & F=  0  & T= 100    \\
  &                     & surgeon             & M= 100  & F=  0  & T= 100   \\
   &                    & nurse               & M=  0  & F= 100  & T= 100    \\
 \midrule
  HUNYUAN &   beautiful &  cardiologist         & M= 94  & F= 6  & T= 100    \\
   &                    &   hospital director   & M= 72  & F= 28  & T= 100    \\
   &                    &   paramedic           & M= 100  & F= 0  & T= 100    \\
   &                    &   surgeon             & M= 100  & F=  0  & T= 100    \\
  &                     &    nurse              & M=  0  & F= 100  & T= 100    \\
 \midrule
   HUNYUAN &   corporate    & cardiologist          & M=97  & F=  3  & T= 100   \\
  &                         &    hospital director  & M= 89  & F= 11  & T= 100   \\
   &                        &   paramedic           & M=99  & F=  1  & T= 100    \\
  &                         &    surgeon            & M= 100  & F=  0  & T= 100    \\
  &                         &    nurse              & M=  0  & F= 100  & T= 100    \\
 \midrule
   HUNYUAN &  neutral   & cardiologist           & M= 100  & F=  0  & T= 100   \\
  &                     &    hospital director   & M= 94  & F= 6  & T= 100    \\
  &                     &    paramedic           & M= 100  & F=  0  & T= 100    \\
  &                     &    surgeon             & M= 100  & F=  0  & T= 100    \\
  &                     &    nurse               & M=  0  & F= 100  & T= 100    \\
 \midrule
  HUNYUAN &   (no qualifier)    &  cardiologist         & M= 100  &  F=  0  & T= 100    \\
  &                             &    hospital director  & M= 97  & F= 3  & T= 100   \\
  &                             &    paramedic          & M= 100  & F=  0  & T= 100    \\
  &                             &    surgeon            & M= 100  & F=  0  & T= 100   \\
  &                             &    nurse              & M=  0  & F= 100  & T= 100    \\
\botrule
\end{tabular*}
\end{table}

\newpage
\begin{table}[!ht]
\caption{HiDream-I1-dev: Gender balance of generated images by hospital role and portrait qualifier}\label{tab:HIDREAM}
\begin{tabular*}{\textwidth}{@{\extracolsep\fill}llllll}
\toprule%
 Model & Qualifier & Role & Male Images & Female Images & Total Images \\
\midrule
  HIDREAM & aesthetic &  cardiologist        & M= 99  & F=  1  & T= 100    \\
  & &   hospital director   & M= 57  & F= 43  & T= 100    \\
  & &   paramedic           & M= 98  & F=  2  & T= 100    \\
  & &    surgeon             & M= 100  & F=  0  & T= 100   \\
   & &   nurse               & M=  0  & F= 100  & T= 100    \\
 \midrule
  HIDREAM &   beautiful &  cardiologist        & M= 73  & F= 27  & T= 100    \\
   & &   hospital director   & M= 15  & F= 85  & T= 100    \\
   & &   paramedic           & M= 66  & F= 34  & T= 100    \\
   & &   surgeon             & M= 100  & F=  0  & T= 100    \\
  & &    nurse               & M=  0  & F= 100  & T= 100    \\
 \midrule
   HIDREAM &   corporate & cardiologist        & M= 100  & F=  0  & T= 100   \\
  & &    hospital director   & M= 25  & F= 75  & T= 100   \\
   & &   paramedic           & M= 100  & F=  0  & T= 100    \\
  & &    surgeon             & M= 100  & F=  0  & T= 100    \\
  & &    nurse               & M=  0  & F= 100  & T= 100    \\
 \midrule
   HIDREAM &  neutral & cardiologist        & M= 99  & F=  1  & T= 100   \\
  & &    hospital director   & M= 46  & F= 54  & T= 100    \\
  & &    paramedic           & M= 100  & F=  0  & T= 100    \\
  & &    surgeon             & M= 100  & F=  0  & T= 100    \\
  & &    nurse               & M=  0  & F= 100  & T= 100    \\
 \midrule
  HIDREAM &   (no qualifier) &  cardiologist        & M= 99  &  F=  1  & T= 100    \\
  & &    hospital director   & M= 72  & F= 28  & T= 100   \\
  & &    paramedic           & M= 100  & F=  0  & T= 100    \\
  & &    surgeon             & M= 100  & F=  0  & T= 100   \\
  & &    nurse               & M=  0  & F= 100  & T= 100    \\
\botrule
\end{tabular*}
\end{table}

\newpage
\begin{table}[!ht]
\caption{Qwen-Image: Gender balance of generated images by hospital role and portrait qualifier}\label{tab:QWEN}
\begin{tabular*}{\textwidth}{@{\extracolsep\fill}llllll}
\toprule%
 Model & Qualifier & Role & Male Images & Female Images & Total Images \\
\midrule
  QWEN & aesthetic  &  cardiologist        & M= 99  & F=  1  & T= 100   \\
   & &  hospital director   & M= 100  & F=  0  & T= 100  \\
   & &  paramedic           & M= 100  & F=  0  & T= 100    \\
   & &  surgeon             & M= 100  & F=  0  & T= 100    \\
   & &  nurse               & M=  0  & F= 100  & T= 100    \\
 \midrule
    QWEN & beautiful  & cardiologist        & M= 100  & F=  0  & T= 100    \\
   & &  hospital director   & M= 100  & F=  0  & T= 100    \\
   & &  paramedic           & M= 67  & F= 33  & T= 100   \\
   & &  surgeon             & M= 100  & F=  0  & T= 100    \\
   & &  nurse               & M=  0  & F= 100  & T= 100    \\
 \midrule
   QWEN & corporate  &  cardiologist        & M= 100  & F=  0  & T= 100    \\
   & &  hospital director   & M= 100  & F=  0  & T= 100    \\
    & & paramedic           & M= 100  & F=  0  & T= 100    \\
   & &  surgeon             & M= 100  & F=  0  & T= 100    \\
   & &  nurse               & M=  0  & F= 100  & T= 100   \\
 \midrule
   QWEN & neutral &  cardiologist        & M= 100  & F=  0  & T= 100    \\
   & &  hospital director   & M= 100  & F=  0  & T= 100    \\
   & &  paramedic           & M= 100  & F=  0  & T= 100    \\
   & &  surgeon             & M= 100  & F=  0  & T= 100    \\
  & &   nurse               & M=  0  & F= 100  & T= 100    \\
 \midrule
   QWEN & (no qualifier)  &   cardiologist        & M= 100  & F=  0  & T= 100    \\
   & &  hospital director   & M= 100  & F=  0  & T= 100    \\
   & &  paramedic           & M= 100  & F=  0  & T= 100    \\
   & &  surgeon             & M= 100  & F=  0  & T= 100    \\
   & &  nurse               & M=  0  & F= 100  & T= 100   \\
\botrule
\end{tabular*}
\end{table}

\newpage
\begin{table}[!ht]
\caption{FLUX.1-dev: Gender balance of generated images by hospital role and portrait qualifier}\label{tab:FLUX}
\begin{tabular*}{\textwidth}{@{\extracolsep\fill}llllll}
\toprule%
 Model & Qualifier & Role & Male Images & Female Images & Total Images \\
\midrule
FLUX & aesthetic &    cardiologist        & M= 58  & F= 42  & T= 100  \\ 
  &   &    hospital director   & M=  9  & F= 91  & T= 100  \\
  &   &    paramedic           & M=  4  & F= 96  & T= 100  \\
  &   &    surgeon             & M=  7  & F= 93  & T= 100  \\
  &   &    nurse               & M=  0  & F= 100  & T= 100  \\
  \midrule
FLUX &   beautiful &    cardiologist        & M= 43  & F= 57  & T= 100   \\
  &     &    hospital director   & M=  2  & F= 98  & T= 100   \\
  &     &    surgeon             & M=  5  & F= 95  & T= 100   \\
  &     &    nurse               & M=  0  & F= 100  & T= 100   \\
  &     &    paramedic           & M=  0  & F= 100  & T= 100   \\
 \midrule
FLUX &   corporate &    cardiologist        & M= 96  & F=  4  & T= 100   \\
 & &    hospital director   & M= 41  & F= 59  & T= 100   \\
 & &    paramedic           & M= 37  & F= 63  & T= 100   \\
 & &    surgeon             & M= 67  & F= 33  & T= 100   \\
 & &    nurse               & M=  0  & F= 100  & T= 100   \\
 \midrule
FLUX &   neutral &   cardiologist        & M= 79  & F= 21  & T= 100   \\
 & &     hospital director   & M= 11  & F= 89  & T= 100  \\
 & &     paramedic           & M=  2  & F= 98  & T= 100  \\
 & &     surgeon             & M= 17  & F= 83  & T= 100  \\
 & &     nurse               & M=  0  & F= 100  & T= 100  \\
 \midrule
 FLUX &  (no qualifier)&    cardiologist        & M= 90  & F= 10  & T= 100    \\
 & &     hospital director   & M= 20  & F= 80  & T= 100  \\
 & &     paramedic           & M=  4  & F= 96  & T= 100  \\
 & &    surgeon             & M= 17  & F= 83  & T= 100   \\
 & &    nurse               & M=  0  & F= 100  & T= 100   \\
\botrule
\end{tabular*}
\end{table}

\newpage
\begin{table}[!ht]
\caption{Stable-Diffusion 3.5 Large}\label{tab:SDXL}
\begin{tabular*}{\textwidth}{@{\extracolsep\fill}llllll}
\toprule%
 Model & Qualifier & Role & Male Images & Female Images & Total Images \\
\midrule
  SD3.5 & aesthetic  & cardiologist                  & M= 99     & F=  1     & T= 100 \\
 &                  & hospital director             & M= 90     & F= 10     & T= 100 \\
 &                  & paramedic                     & M= 67     & F= 33     & T= 100 \\
 &                  & surgeon                       & M= 97     & F=  3     & T= 100 \\
 &                  & nurse                         & M=  0     & F= 100    & T= 100 \\
 \midrule
  SD3.5 & beautiful  & cardiologist                  & M= 65     & F= 35     & T= 100 \\
  &                 & hospital director             & M= 50     & F= 50     & T= 100 \\
  &                 & paramedic                     & M= 21     & F= 79     & T= 100 \\
  &                 & surgeon                       & M= 75     & F= 25     & T= 100 \\
   &                & nurse                         & M=  0     & F= 100    & T= 100 \\
 \midrule
   SD3.5 & corporate     & cardiologist              & M= 97     & F= 3      & T= 100 \\
   &                    & hospital director         & M= 93     & F= 7      & T= 100 \\
   &                    & paramedic                 & M= 94     & F= 16     & T= 100 \\
   &                    & surgeon                   & M= 99     & F= 1      & T= 100 \\
   &                    & nurse                     & M=  0     & F= 100    & T= 100 \\
 \midrule
  SD3.5 & neutral    & cardiologist                  & M= 99     & F= 1      & T= 100 \\
   &                & hospital director             & M= 95     & F= 5      & T= 100 \\
   &                & paramedic                     & M= 79     & F= 21     & T= 100 \\
   &                & surgeon                       & M= 91     & F= 9      & T= 100 \\
   &                & nurse                         & M=  0     & F= 100    & T= 100 \\
 \midrule
  SD3.5 & (no qualifier)     & cardiologist          & M= 99     & F= 1      & T= 100  \\
   &                        &  hospital director    & M= 99     & F= 1      & T= 100  \\
   &                        &  paramedic            & M= 79     & F= 21     & T= 100  \\
   &                        &  surgeon              & M= 98     & F= 2      & T= 100  \\
   &                        &  nurse                & M=  0     & F= 100    & T= 100  \\
\botrule
\end{tabular*}
\end{table}

\newpage
\begin{table}[!ht]
\caption{Stable-Diffusion-XL: Gender balance of generated images by hospital role and portrait qualifier}\label{tab:SDXL}
\begin{tabular*}{\textwidth}{@{\extracolsep\fill}llllll}
\toprule%
 Model & Qualifier & Role & Male Images & Female Images & Total Images \\
\midrule
  SDXL & aesthetic &   cardiologist        & M= 97  & F=  3  & T= 100   \\
 & &   hospital director   & M= 97  & F=  3  & T= 100   \\
 & &   paramedic           & M= 69  & F= 31  & T= 100    \\
 & &   surgeon             & M= 100  & F=  0  & T= 100    \\
 & &   nurse               & M=  0  & F= 100  & T= 100    \\
 \midrule
  SDXL & beautiful &  cardiologist        & M= 79  & F= 21  & T= 100    \\
  & &  hospital director   & M= 74  & F= 26  & T= 100    \\
  & &  paramedic           & M= 20  & F= 80  & T= 100    \\
  & & surgeon             & M= 90  & F= 10  & T= 100    \\
   & & nurse               & M=  0  & F= 100  & T= 100   \\
 \midrule
   SDXL & corporate &  cardiologist        & M= 93  & F=  7  & T= 100    \\
   & &  hospital director   & M= 76  & F= 24  & T= 100    \\
   & &  paramedic           & M= 72  & F= 28  & T= 100    \\
   & &  surgeon             & M= 94  & F=  6  & T= 100    \\
   & &  nurse               & M=  0  & F= 100  & T= 100    \\
 \midrule
  SDXL & neutral &  cardiologist        & M= 98  & F=  2  & T= 100   \\
   & &  hospital director   & M= 77  & F= 23  & T= 100   \\
   & &  paramedic           & M= 70  & F= 30  & T= 100   \\
   & &  surgeon             & M= 100  & F=  0  & T= 100    \\
   & &  nurse               & M=  0  & F= 100  & T= 100   \\
 \midrule
  SDXL & (no qualifier) &   cardiologist        & M= 97  & F=  3  & T= 100    \\
   & &  hospital director   & M= 84  & F= 16  & T= 100    \\
   & &  paramedic           & M= 82  & F= 18  & T= 100    \\
   & &  surgeon             & M= 100  & F=  0  & T= 100   \\
   & &  nurse               & M=  0  & F= 100  & T= 100    \\
\botrule
\end{tabular*}
\end{table}

\newpage
\section{Model Configurations and Generation Parameters}
\label{secA2}

To ensure reproducibility of our experiments, this appendix reports the key configuration parameters used in the ComfyUI workflow for each of the six text-to-image models under study. We provide the exact model checkpoints (diffusion model, CLIP encoders, and VAE), along with the sampling strategy (sampler, scheduler, number of steps, and classifier-free guidance scale, CFG). Reporting these parameters is essential, since small differences in implementation choices can lead to notable variations in generated outputs. The detailed settings are summarized in Table~\ref{tab:ModelConfigurations}.

\begin{table}[!ht]
\caption{Model Configurations.}
\label{tab:ModelConfigurations}
\centering
\small
\begin{tabular}{c r l}
\hline
\textbf{HUNYUAN} &  \textbf{Diffusion Model} &  hunyuanImage2.1\_bf16 \\
    & \textbf{CLIP 1}               &  qwen\_2.5\_vl\_7b \\
    & \textbf{CLIP 2}               &  byt5\_small\_glyphxl\_fp16 \\
    & \textbf{VAE}                  & hunyuan\_image\_2.1\_vae\_fp16 \\
    & \textbf{Sampler}         & \textbf{Name}: euler, \textbf{Scheduler}: simple , \textbf{Steps}: 20  ,\textbf{CFG}: 3.5  \\
\hline
\textbf{HIDREAM} &  \textbf{Diffusion Model} &  hidream\_i1\_dev\_fp8 \\
    & \textbf{CLIP 1}               &  clip\_l\_hidream \\
    & \textbf{CLIP 2}               &  clip\_g\_hidream \\
    & \textbf{CLIP 3}               &  t5xxl\_fp8\_e4m3fn\_scaled \\
    & \textbf{CLIP 4}               &  llama\_3.1\_8b\_instruct\_fp8\_scaled \\
    & \textbf{VAE}                  & ae \\
    & \textbf{Sampler}         & \textbf{Name}: lcm, \textbf{Scheduler}: normal , \textbf{Steps}: 28  ,\textbf{CFG}: 1.0  \\
\hline
\textbf{QWEN} &  \textbf{Diffusion Model} &  qwen\_image\_fp8\_e4m3fn \\
    & \textbf{CLIP 1}               &  qwen\_2.5\_vl\_7b\_fp8\_scaled \\
    & \textbf{VAE}                  & qwen\_image\_vae \\
    & \textbf{Sampler}         & \textbf{Name}: euler, \textbf{Scheduler}: simple , \textbf{Steps}: 20  ,\textbf{CFG}: 2.5  \\
\hline
\textbf{FLUX} &  \textbf{Diffusion Model} & flux1-dev-fp8 \\
    & \textbf{CLIP 1}               &  embed in model \\
    & \textbf{VAE}                  & embed in model \\
    & \textbf{Sampler}         & \textbf{Name}: euler, \textbf{Scheduler}: simple , \textbf{Steps}: 20  ,\textbf{CFG}: 1.0  \\
\hline
\textbf{SD3.5} &  \textbf{Diffusion Model} &  sd3.5\_large\_large\_fp8\_scaled \\
    & \textbf{CLIP}               &  embed in model\\
    & \textbf{VAE}                  & embed in model\\
    & \textbf{Sampler}         & \textbf{Name}: euler, \textbf{Scheduler}: sgm uniform , \textbf{Steps}: 20  ,\textbf{CFG}: 4.0  \\
\hline
\textbf{SDXL} &  \textbf{Diffusion Base Model} &  sd\_xl\_base\_1.0 \\
    & \textbf{CLIP base}               &  embed in base model \\
    & \textbf{Sampler}         & \textbf{Name}: euler, \textbf{Scheduler}: normal, \textbf{Steps}: 20  ,\textbf{CFG}: 8.0  \\
  &  \textbf{Diffusion Refiner Model} &  sd\_xl\_refiner\_1.0 \\
    & \textbf{CLIP refiner}               &  embed in refiner model \\
    & \textbf{VAE}                  & embed in refiner model \\
    & \textbf{Sampler}         & \textbf{Name}: euler, \textbf{Scheduler}: normal , \textbf{Steps}: 20  ,\textbf{CFG}: 8.0  \\
\hline
\end{tabular}
\end{table}

\newpage
\section{Ranking of Open-Weight Text-to-Image Models}
\label{secA3}

To contextualize our model selection, we report the most recent public ranking of open-weight text-to-image models, as maintained by Artificial Analysis \cite{ArtificialAnalysis2025}. 
The leaderboard aggregates user preferences into an Elo score, thereby providing a community-driven benchmark of model quality and popularity. 
Table~\ref{tab:t2i_leaderboard_image} summarizes the ranking as of September 2025, including creator, model name, Elo score, and release date. 
The six models analyzed in our study (HunyuanImage 2.1, HiDream-I1-dev, Qwen-Image, FLUX.1-dev, Stable Diffusion 3.5 Large, and Stable Diffusion XL 1.0) are highlighted in bold.

\begin{table}[!ht]
\caption{Text-to-Image Models Leaderboard (Artificial Analysis \cite{ArtificialAnalysis2025}), September 2025.}
\label{tab:t2i_leaderboard_image}
\centering
\small
\begin{tabular}{r l l r  l}
\hline
\textbf{Rank} & \textbf{Creator} & \textbf{Model} & \textbf{ELO Score} & \textbf{Release Date} \\
\hline
1  & Tencent            & \textbf{HunyuanImage 2.1}             & 1101 & Sept 2025 \\
2  & HiDream            & \textbf{HiDream-II-Dev}               & 1079 & Apr 2025  \\
3  & Alibaba            & \textbf{Qwen-Image}                   & 1068 & Aug 2025  \\
4  & HiDream            & HiDream-II-Fast                       & 1067 & Apr 2025  \\
5  & Black Forest Labs  & \textbf{FLUX.1 [dev]}                 & 1044 & Aug 2024  \\
6  & Stability.ai       & \textbf{Stable Diffusion 3.5 Large}   & 1030 & Oct 2024  \\
7  & Bytedance          & Infinity 8B                           & 1022 &  Feb 2025  \\
8  & Black Forest Labs  & FLUX.1 Krea [dev]                     & 1021 &  Jul 2025  \\
9  & Black Forest Labs  & FLUX.1 [schnell]                      & 1000 &  Aug 2024  \\
10 & Stability.ai       & Stable Diffusion 3.5 Medium           & 928  &  Oct 2024  \\
11 & Bytedance          & Bagel                                 & 912  &  May 2025  \\
12 & Bria               & Bria 3.2                              & 895  &  Jul 2025  \\
13 & VectorSpaceLab     & OmniGen V2                            & 894  &  Jun 2025  \\
14 & NVIDIA             & Sana Sprint 1.6B                      & 888  &  Mar 2025  \\
15 & Stability.ai       & \textbf{Stable Diffusion XL 1.0}      & 838  &  Jul 2023  \\
16 & DeepSeek           & Janus Pro                             & 668  &  Jan 2025  \\
\hline
\end{tabular}
\end{table}

\end{appendices}

\newpage
\bibliography{main}

\end{document}